\pdfoutput=1

\documentclass[acmsmall, review=false, screen]{acmart}
\usepackage[subrefformat=parens,labelformat=parens]{subfig}
  \usepackage{xspace} 
\usepackage{pifont}
\usepackage{balance}
\usepackage{diagbox}
\usepackage{paralist} 
\usepackage{multirow} 
\usepackage{rotating} 
\usepackage{makecell} 
\usepackage{enumitem} 

\newsavebox\CBox

\newtheorem{innercustomthm}{Observation}
\newenvironment{customthm}[1]
{\renewcommand\theinnercustomthm{#1}\innercustomthm}
{\endinnercustomthm}

\newcommand{\bit}{\begin{itemize}}
	\newcommand{\eit}{\end{itemize}}

\theoremstyle{definition}
\newtheorem{definition}{Definition}[section]

\newcommand{\hide}[1]{}

\newcommand{\kreg}{{$k$-regular\xspace}}

\newcounter{x}

\newcommand{\mcG}{{G}}
\newcommand{\mcV}{\mathcal{V}}
\newcommand{\mcE}{\mathcal{E}}
\newcommand{\mcK}{\mathcal{K}}
\newcommand{\mcH}{\mathcal{H}}
\newcommand{\bX}{\mathbf{X}}
\newcommand{\bx}{\mathbf{x}}
\newcommand{\R}{\mathbb{R}}

\newcommand{\rad}{{\it NN-Radius}\xspace}
\newcommand{\dis}{{\it NN-Disagreement\%}\xspace}

\newcommand{\flip}{\cellcolor{yellow!20}}
\newcommand{\good}{\cellcolor{green!20}}
\newcommand{\bad}{\cellcolor{red!20}}

\newcommand{\projurl}{\url{https://github.com/LingxiaoShawn/GLOD-Issues}}

\newcommand{\BFSERIES}{\fontseries{b}\selectfont}

\acmISBN{}
\acmDOI{}
\acmPrice{}
\acmConference[]{}{}{}
\acmBooktitle{}
\begin{document}

\title{On Using  Classification Datasets to Evaluate Graph Outlier Detection: 
	Peculiar Observations and New Insights}

\author{Lingxiao Zhao}
\email{lingxia1@andrew.cmu.edu}
\affiliation{%
	\institution{Heinz College - Information Systems \& Public Policy, Carnegie Mellon University}
	\streetaddress{5000 Forbes Ave}
	\city{Pittsburgh}
	\state{PA}
	\country{USA}
	\postcode{15213}
}
\author{Leman Akoglu}
\email{lakoglu@andrew.cmu.edu}
\affiliation{%
 \institution{Heinz College - Information Systems \& Public Policy, Carnegie Mellon University}
	\streetaddress{5000 Forbes Ave}
	\city{Pittsburgh}
	\state{PA}
	\country{USA}
	\postcode{15213}
}

\thanks{}

\begin{abstract}


It is common practice of the outlier mining community to repurpose classification datasets
toward evaluating various detection models.
To that end, often a binary classification dataset is used, where samples from 
one of the classes is designated as the `inlier' samples, and the other class is substantially down-sampled to create the (ground-truth) `outlier' samples. Graph-level outlier detection (GLOD) is rarely studied but has many potentially influential real-world applications. 
In this study, we identify an intriguing issue with repurposing graph classification datasets for \textbf{GLOD}. We find that ROC-AUC performance of the models changes significantly (``flips'' from high to very low, even worse than random) depending on which class is down-sampled. Interestingly, ROC-AUCs on these two variants approximately sum to 1 and their performance gap is amplified with increasing propagations for a certain family of {\em propagation based} outlier detection models.
We carefully study the graph embedding space produced by propagation based models and find two driving factors: (1) disparity between within-class densities which is amplified by propagation, and (2) overlapping support (mixing of embeddings) across classes. 
We also study other graph embedding methods and downstream outlier detectors, and find that the intriguing ``performance flip'' issue still widely exists but which version of the downsample achieves higher performance may vary. Thoughtful analysis over comprehensive results further deeper our understanding of the established issue.


With this study, we aim to draw attention to this (to our knowledge) previously-unnoticed issue for the rarely studied GLOD problem, and specifically to the following questions: 1) Given the performance flip issue we identified, where one version of the downsample often yields worse-than-random performance, is it acppropriate to evaluate GLOD by average performance across all downsampled versions when repurposing graph classification datasets? 2) Considering onsistently observed performance flip issue across different graph embedding methods we studied, is it possible to design better graph embedding methods to to overcome the issue? We conclude the paper with our insights to these questions.


\end{abstract}
\maketitle 

\section{Introduction}
\label{sec:introduction}

Outlier detection is a critical task that finds numerous applications in healthcare, security, finance, etc. \cite{aggarwal2015outlier}.
Simply put, the task is to identify observations that notably stand out within large collections of data so as to ``arouse suspicions that [they were] generated by a different mechanism'' \cite{hawkins1980identification}.
One of the key challenges of outlier detection is that it poses an \textit{unsupervised} learning problem.
Due to the rare nature of outlier instances, combined with the laborious manual (i.e., human) labeling, 
access to benchmark datasets with sufficiently many labeled ground-truth outliers is limited.

{\bf Motivation.~} Lack of labeled benchmark datasets for outlier detection is not only a challenge for learning, but also for the \textit{evaluation} of outlier models.
Even if one designs unsupervised models for the detection task, ground-truth labels that truly reflect the nature of outliers in a domain is essential  for the reliable error estimation of various models.
Thereby, the scarcity of representative labeled outliers in real-world datasets has motivated a couple of strategies for building benchmark datasets, mainly for evaluation. 

One strategy is to inject realistic yet synthetic outliers into real-world datasets via simulation. For example, in fraud detection applications, one could simulate activities that reflect malicious schemes known to domain experts in order to obtain positive (i.e., anomalous) observations.
Emmott \textit{et al.} \cite{emmott2013systematic,emmott2015meta} present a systematic in-depth study on this subject.
This approach is typically criticized for a couple of reasons. First, the simulated outliers are limited to the known anomalous behaviors and may not comprehensively reflect the outliers in the wild. Second, this type of approach 
may create an environment fertile to ``leakage'', where the outliers may be simulated in a biased way that aligns with how the detection model under evaluation works. 

An alternative strategy to artificial outlier injection is to repurpose classification datasets so as to work  with only real-world samples. (See \cite{campos2016evaluation} and citations therein.)
  A common practice is to use binary classification datasets, where samples from one of the classes (typically the one with the larger number of samples) is treated as the `inlier' samples, and the other class is down-sampled (to a desired rate) to constitute the `outlier' samples.
  This procedure conforms with the notion of outlierness as characterized by Hawkins \cite{hawkins1980identification}, in that the outliers are drawn from a data distribution (i.e., class) that is different from that generating the inliers.
  In their in-depth evaluation of unsupervised outlier detection models, Campos \textit{et al.} 
  \cite{campos2016evaluation} mainly adopt this strategy. 
  


{\bf This paper.~} In this study we scrutinize this latter strategy, and pose the following questions: \textit{Should one use classification datasets for evaluating outlier detection models? What issues should one be aware of in designing benchmark datasets in this manner?}
Specifically, we study this issue in the context of outlier detection in \textit{graph databases}, where given a collection of graphs, the task is to identify the outlier graphs that stand out.
 Graph data is widespread in finance, health care, cybersecurity, fault monitoring, etc. where the outlier detection task finds a long list of applications such as identifying rare transaction graphs \cite{nguyen2020anomaly}, command flow graphs \cite{conf/kdd/ManzoorMA16}, and human poses \cite{markovitz2020graph}, fake news \cite{monti2019fake}, traffic events \cite{harshaw2016graphprints}, buggy software  \cite{liu2005mining}, money laundering \cite{weber2019antimoney}, and so on.



%


Before delving into details, we start by illustrating the intriguing ``performance flip'' issue empirically. Table \ref{tab:flip} (See Sec. \ref{ssec:obs}) shows the ROC-AUC performances of three graph embedding based outlier detectors based on four binary graph classification datasets (Additional results on more datasets, and using more graph embedding methods is available in Tables \ref{tab:all_xnonx}\&\ref{tab:all_xy} ). Each dataset has two variants, each corresponding to one of the classes down-sampled as outlier.
The difference in performances between the two variants is striking, consistently across models on most (although not all) datasets.

{\bf Related work.~} To the best of our knowledge, the performance flip issue has not been identified by any prior work on outlier mining, with the exception of work by Swersky \textit{et al.}  \cite{swersky2016evaluation} which document similar ROC-AUC flip behavior on several datasets, however the authors have not recognized explicitly.
Campos \textit{et al.} \cite{campos2016evaluation} state that ``random downsampling often leads to great variation in the nature of the outliers produced'' and that  ``observations based on downsampling can vary considerably from sample to sample'', based on which they repeat their down-sampling procedure 10 times per dataset ``to mitigate the impact of randomization''. This, however, points to an orthogonal issue as it pertains to down-sampling {\em after} deciding (i.e. fixing) \textit{which} class to down-sample. 
Repurposing classification datasets for evaluation of clustering has been questioned by F{\"a}rber \textit{et al.} \cite{farber2010using}, which alludes to the potential misalignment between the semantics of data clusters and class labels.  Our study points to an issue orthogonal to semantics.


{\bf Contributions.~} Through extensive analysis, our study aims to 
(1) illustrate the issues with using graph classification datasets for creating outlier benchmarks for model evaluation,
(2) identify the leading factors behind these issues,
(3) propose concrete measures to quantify these factors and explain their possible driving mechanisms with a focus on propagation-based graph embedding methods,
(4) analyze the root of the issue from three different perspectives (data, graph embedding method, assumption of outlier detector) and call community's attention to three important questions
regarding ($i$) fair evaluation, ($ii$) model selection, and ($iii$) suitability of graph embedding method,
and last but not the least 
(5) open source all methods and datasets used in our study (\projurl), to enable the community to use in their ``GLOD'' tasks and also to facilitate further investigation into the issues raised through our study.
We summarize our main contributions as follows.
\bit
\item {\bf Study of Deep Graph-level Outlier Detection:}~ We start with the design and evaluation of two different categories of  
models for outlier detection  in graph databases; namely, (1) two-stage models---pipelining unsupervised graph-level representation learning with off-the-shelf point-cloud outlier detectors, and (2) end-to-end models---learning representations simultaneously with optimizing an anomaly detection objective, such as one-class classification or reconstruction loss. (Sec. \ref{sec:prelim})

\item {\bf  ``Performance Flip'' Issue with Using Classification Datasets for Evaluation:}~ 
To evaluate the aforementioned graph outlier models, we construct labeled benchmark datasets by repurposing binary graph classification datasets.
Notably, we 
 down-sample those datasets in \textit{both} ``directions'', that is, we create \textit{two} benchmark variants \textit{per} classification dataset, respectively down-sampling one or the other class samples to constitute the outlier graphs.
 Surprisingly, we find that most models, while achieving high detection performance on one variant, fail considerably on the other. That is, we identify the intriguing issue of what we call ``performance flip'' depending on which class has been down-sampled.  (Sec. \ref{ssec:obs})

\item {\bf 
	Driving Factors behind ``Performance Flip'':}~ We find that the issue stems from the (mis)alignment between the inlier/outlier distributions created by graph embedding techniques and key underlying assumptions of the detection models. While one scenario creates a dense inlier distribution surrounded with dispersed outliers (`easy' task), the other creates a sparse inlier distribution that has overlapping support with a small set of outliers with relatively higher density (`hard' task). Since most models assume the former scenario in their formalism, they `do well' on the respective `easy' task. 

{\hspace{.3cm}} With an in-depth study over propagation based graph embedding methods, we identify two key leading factors behind the observed ``performance flip'' issue, particularly (1) \textit{density disparity}; where the density of graph embeddings differ considerably between two classes, and  (2) \textit{overlapping support}; where the distributions of graph embeddings from the two classes exhibit overlapping support in the representation space.
Moreover, we point out two contributing factors: 
(a) initial disparity between within-class sample similarities, and (b) amplification of this disparity by graph propagation -- called \textit{sparsification} -- which is a property of some graph embedding models (Sec \ref{ssec:hypothesis}).
We design quantitative metrics to concretely measure those factors (Sec. \ref{ssec:measures}), and analyze the sparsification property via controlled simulations on $k$-regular graphs (Sec. \ref{ssec:simulate}). Finally, we
present a detailed empirical study on real-world datasets (Sec. \ref{sec:exp}).
We also present additional results on more datasets, using other detector and embedding methods in Sec. \ref{sec:additional_result}. Several additional observations are summarized which deepen our understanding of performance flip.

\item {\bf Insights for Graph-Level Outlier Detection and Beyond:
}~ 
The performance flip issue is also observed for various other embedding methods and  outlier detectors beyond propagation based methods, but which version of the downsample has higher ROC-AUC varies. The persistence of performance flip, but inconsistence of which version achieves higher ROC-AUC, raise several important problems to tackle GLOD: (1) as the performance flip is widely observed for all embedding methods, simply averaging performance among two versions of downsample seem problematic, as one version often has worse-than-random performance; (2) given the challenge of model selection for unsupervised methods, choosing which method to use is not only hard but also becomes risky as one may suffer from the worse-than-random performance; (3) as embedding methods play a large impact on outcomes, a better solution may involve designing an unsupervised graph embedding method that can generate clustered embeddings for different classes, which appears to be a hard problem for unsupervised tasks like outlier detection.


{\hspace{.3cm}} We also argue that issues we identify may extend beyond outlier detection, with possible implications on graph classification and clustering.
Given the popularity of graph neural networks (GNNs), we point out that almost all GNN models employ a message-passing based propagation mechanism, as such, they also have the potential of suffering from sparsification. 
Specifically, this can cause severe overfitting for graph classification when the number of labeled samples is small.
It can also adversely affect graph-level clustering tasks that aim to identify dense regions in the (representation) space.
(Sec. \ref{sec:discuss})

\eit



%
\section{Key Background:  Problem \& Outlier Models}
\label{sec:prelim}

In this paper we focus on the graph-level outlier detection problem. 
The intriguing ``performance flip'' issue we observe 
arises from repurposing binary graph classification datasets for outlier detection evaluation. 
As far as we know, there is limited work studying the graph-level outlier detection problem, where the goal is to discover graphs with rare, unusual patterns which can be distinguished from the majority of graphs in a database. We call attention to the problem as it applies to many important real-world tasks from diverse domains such as drug discovery, money laundering, molecular synthesis, rare human pose detection \cite{markovitz2020graph}, fake news detection \cite{monti2019fake}, traffic events detection \cite{harshaw2016graphprints}, and buggy software detection \cite{liu2005mining}.


\subsection{Graph-Level Outlier Detection}
Let $G = (\mcV,\mcE, \bX)$ be an attributed or labeled graph with $\mcV$ and $\mcE$ depicting its 
vertex set and edge set, where each node $i\in \mcV$ is associated with a feature vector $\bx_i \in \R^d$ and $\bX = [\bx_1, \ldots, \bx_n]^T$ denotes the feature matrix, $n=|\mcV|$ being the total number of nodes. For labeled graphs, each node feature vector $\bx_i \in \R^d$ is a one-hot encoded vector with $d$ being the total number of unique (discrete) node labels. 
\begin{definition}[Graph-Level Outlier Detection Problem (GLOD)]
Given a graph database $\mathcal{G}=\{G_1, \ldots, G_N\}$ containing $N$ labeled or attributed graphs, find the graphs that differ significantly from the majority of graphs in $\mathcal{G}$.
\end{definition}
The above problem is a general statement for graph-level outlier detection. In the real-world how one defines rareness or the degree of difference to the majority may be critical and may change depending on the application. 

\subsection{Graph-level Outlier Detection Models}
\label{ssec:models}
Although there is no specifically designed method existing for GLOD, 
several methods for solving graph classification can be easily modified for tackling the problem. In this paper we mainly focus 
on three \textbf{propagation} based methods that can be categorized as two types: two-stage versus end-to-end. For the purposes of this paper, we find these three methods to be sufficiently illustrative of the issues we discover. \textbf{Notice that we focus on studying propagation based methods because all message-passing based GNNs belong to this category, which are the most promising and popular models for graph representation learning.} In additional experiments presented in Sec. \ref{sec:additional_result}, we also show results for two additional unsupervised graph embedding methods to demonstrate the persistence of the performance flip issue, namely Graph2Vec \cite{narayanan2017graph2vec} and FGSD \cite{verma2017hunt}. Graph2Vec generates graph-level embeddings via Word2Vec by viewing motifs as ``words'' and the graph as a ``document''. FGSD embeds a graph as a histogram of all node spectral distances without using information of node labels. To avoid distraction from the carefully-studied propagation based methods, we omit their details and refer readers to the original papers (\cite{narayanan2017graph2vec,verma2017hunt}). 

\subsubsection{Two-Stage Graph Outlier Detection} \label{sec:dominance}
Two-stage graph outlier detection approaches first transform graphs into graph embeddings or similarities between graphs by using unsupervised graph embedding methods (such as graph2vec \cite{narayanan2017graph2vec} and FGSD \cite{verma2017hunt}) or graph kernels (such as feiler-Leman kernel \cite{shervashidze2011weisfeiler} and propagation kernel \cite{neumann2016propagation}). Then traditional outlier detectors such as Isolation Forest \cite{liu2008isolation}, Local Outlier Factor (LOF \cite{breunig2000lof}), and one-class SVM (OCSVM) \cite{manevitz2001one} can be used to detect outliers in the embedding (vector) space. These approaches are easy to use and do not require much hyperparameter tuning, which makes them relatively stable for an unsupervised task like outlier detection. Nevertheless, two-stage methods may suffer from suboptimal solution as the feature extractor and outlier detector are independent. Moreover, most unsupervised graph feature extractors produce ``hand-crafted'' features that are deterministic without much room to improve, which further restrict the capacity of two-stage methods. 

To illustrate the performance flip issue, we focus on graph kernel based two-stage approaches with two well-known outlier detectors used downstream: OCSVM and LOF. A graph kernel defines a kernel function $\mcK$ that outputs a similarity between two graphs. Formally it can be written as
\begin{align}
	\mcK(G, G^\prime) = \langle \phi(\mcG), \phi(\mcG^\prime)  \rangle_{\mcH}
\end{align}
where $\mcH$ is a RKHS and $\langle \cdot, \cdot \rangle$ is the dot product in $\mcH$. The mapping $\phi(\mcG)$ transforms graph $\mcG$ to an embedding vector in $\mcH$, which in our case contains counts of atomic subgraph patterns.   Specifically we use the Weisfeiler-Leman subtree kernel and the propagation kernel, described as follows.

\noindent
\textbf{Weisfeiler-Leman Subtree Kernel.} 
Inspired by Weisfeiler-Leman (WL) test of graph isomorphism \cite{weisfeiler1968reduction} (a.k.a. the color-refinement algorithm), WL subtree kernel \cite{shervashidze2011weisfeiler} processes a labeled graph by iteratively re-labeling each vertex with a new label compressed from a multiset label consisting of the vertex's original label and the sorted labels of its neighbors. This procedure repeats for $L$ iterations for all graphs and outputs $L$ re-labeled graphs $\{\mcG_1, ..., \mcG_L\}$ for every graph $\mcG$. One can easily show that each vertex in $\mcG_l$ at $l$ iterations represents the subtree of the original vertex with depth $l$. WL subtree kernel compares two graphs by simply counting the number of co-occurrences of labels in both graphs at each iteration. The similarity score of two graphs is the summation of similarities across iterations. Formally, one can write it as 
\begin{align}
	\mcK_{WL} (\mcG, \mcG^\prime) = \sum_{l=0}^L  \langle \phi_{WL}(\mcG_l), \phi_{WL}(\mcG^\prime_l)  \rangle
\end{align}
where $G_0$ represents the original input graph, and  $\phi_{WL}$ counts the frequency of all labels in the input graph by a vector with length equal to the number of unique labels. 

\begin{figure}[h!]
    \centering
    \includegraphics[width=\textwidth]{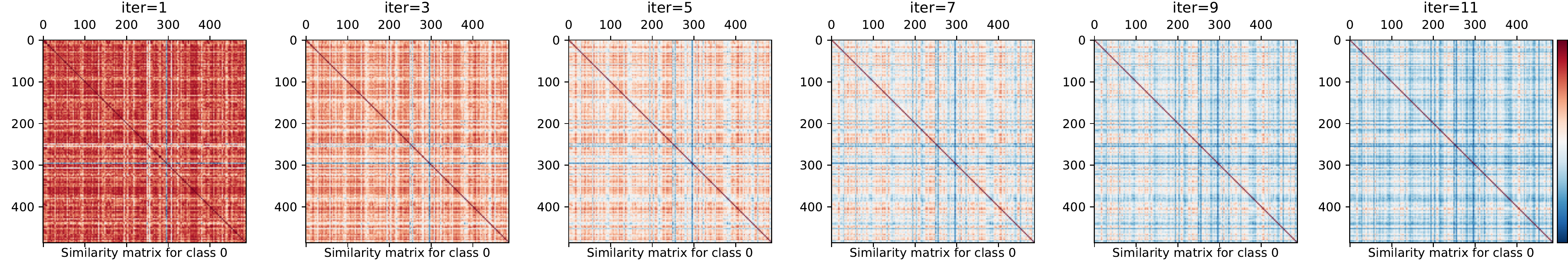}
    \caption{Sparsification in WL subtree kernel: Pairwise similarity of graphs (from DD dataset) decreases with increasing number of iterations (left to right).}\label{fig:dominance-WL}
\end{figure}

\textbf{Sparsification.} Next we highlight a key property of the WL subtree kernel that is closely related to the performance flip issue we discover. 
Substructure-based graph kernels consider each substructure as a separate feature to compare among graphs. The total number of distinct substructures grows exponentially in the diameter of the substructures, which leads to the \textit{sparsity problem} --- that only a limited number of substructures would be shared among graphs. This property has also been referred to as \textit{diagonal dominance} \cite{yanardag2015deep,narayanan2016subgraph2vec} --- wherein 
 each individual graph would mostly be similar only to itself but not much to any other graph. 
 Being based on substructures, WL subtree kernel distinguishes each $k$-hop subgraph as a separate feature (re-labeled as a different label), as such, its feature space tends to grow exponentially in the number of iterations. The sparsification property of WL kernel is visualized in Fig. \ref{fig:dominance-WL}, where the diagonal dominance and diminishing similarity among graphs are observed clearly. 

\noindent
\textbf{Propagation Kernel.} Propagation kernel (PK) \cite{neumann2016propagation} is inspired from the idea of propagating label information among nodes over the graph structure such as label propagation algorithm \cite{citeulike310457} for semi-supervised node classification and can be used for both attributed graphs and one-hot encoded labeled graphs. For each graph $\mcG=(\mcV,\mcE,\bX)$, let $X_0=\bX$ denote the original feature matrix. Then PK generates a new feature matrix at each iteration by propagating the feature matrix using the transition matrix $T=D^{-1}A$ (where $A$ is the adjacency matrix and $D$ is the diagonal degree matrix) of the graph. Formally, 
$\bX_{l+1} = T\bX_{l}$. Similar to WL subtree kernel, PK compares two graphs at each iteration. The similarity between two graphs is measured based on propagated features through binning. Formally we can write the kernel as 
\begin{align}
\label{eq:pk}
	\mcK_{PK} (\mcG, \mcG^\prime) = \sum_{l=0}^L  \langle \phi_{PK}(\bX^{\mcG}_l), \phi_{PK}(\bX^{\mcG^\prime}_l)  \rangle
\end{align} 
where $\phi_{PK}(\cdot)$ denotes the hash function that maps a given set of (feature) vectors into bins. To preserve locality and keep efficiency, locally sensitive hashing (LSH) \cite{citeulike556224} is used for the binning.

\begin{figure}[h!]
    \centering
    \includegraphics[width=\textwidth]{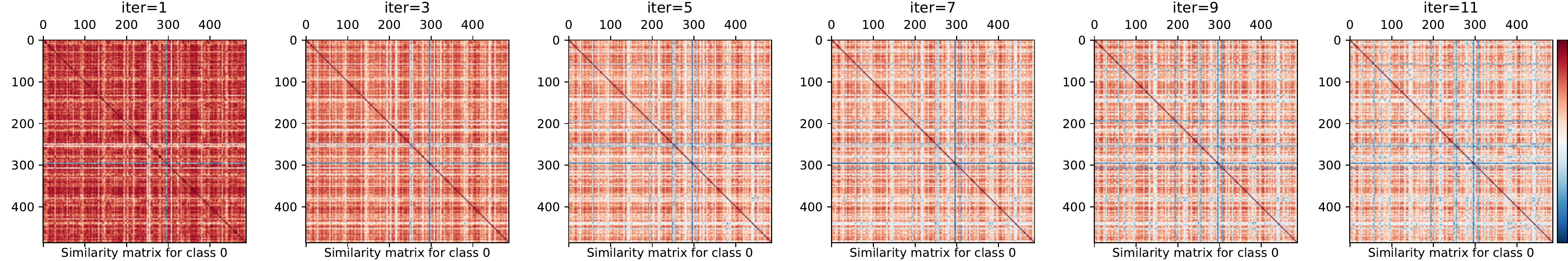}
        \caption{Sparsification in PK: Pairwise similarity of graphs (from DD dataset) decreases with increasing number of iterations (left to right).}\label{fig:dominance-PK}
\end{figure}

\textbf{Sparsification.}  The propagation kernel also exhibits the aforementioned sparsity problem, increasingly for larger number of iterations.
Compared to WL subtree kernel that generates new features via re-labeling (a hard transformation), propagation kernel generates new features via multiplying by the transition matrix (a soft transformation). Thus, the feature space  grows much slower for PK. As illustrated in Fig. \ref{fig:dominance-PK}, the diagonal dominance continues to hold but the sparsification occurs at a lower rate than WL subtree kernel (cf. Fig. \ref{fig:dominance-WL}). 

\subsubsection{End-to-End Deep Graph Outlier Detection}
Deep learning methods have been used for outlier detection recently to enhance automatic feature learning for high-dimensional and structured data such as images. Recently graph neural network (GNN) has achieved great success in graph-structured data, and several works have successfully applied GNNs to node-level outlier detection on a single graph, such as OCGNN \cite{wang2020ocgnn} and DOMINANT \cite{ding2019deep}. However there is no deep model proposed for graph-level outlier detection. 

Here we present a GNN model adapted from graph classification, and leverage a one-class classification objective function to address graph-level outlier detection. Compared with the widely used Graph Convolution Network (GCN) \cite{kipf2017semi} model, Graph Isomorphism Network (GIN) \cite{conf/iclr/XuHLJ19} has been shown to be as powerful as the WL test of graph isomorphism, as such, we design a GIN based graph-level outlier detector. Note that an earlier GNN model called DGCNN \cite{zhang2018end} has discussed its connection to WL subtree kernel and propagation kernel. GIN builds on the ideas of DGCNN, and as a result also shares connection to these graph kernels. As we will present the issues we have discovered based on those graph kernels, we have also empirically verified that similar issues are observed for the GIN based model. In the following, we present our GIN based graph-level outlier detector, which is trained end-to-end through one-class classification loss. 

Let $h_v^{(l)}$ be the $l$-th layer representation of node $v$ in the GIN model. GIN updates node representations at each layer by 
\begin{align}
	h_{v}^{(l)} = \text{MLP}^{(l)} \Big( (1+\epsilon^{(l)}) \cdot h_v^{(l-1)} 
	+ \sum_{u\in \mathcal{N}(v)} h_u^{(l-1)}   \Big)
\end{align}
where MLP denotes a multi-layer perceptron (we use 2 layer) and $\mathcal{N}(v)$ denotes the direct neighbors of node $v$. Note that the summation operation around neighbor vectors can cause numerical explosion over iterations, thus batch normalization is applied between each GIN layer to prevent it. 

After $L$ layers, GIN generates the graph-level representation (i.e. graph embedding) using a readout function as follows.
\begin{align}
	h_{\mcG} = \text{CONCAT}\Big(\; \text{READOUT}(\{h_{v}^{(l)} |v\in \mcG \}) \;|\; l=0,1,\ldots,L \;\Big) \;.
\end{align}
While the original paper \cite{conf/iclr/XuHLJ19} proposed summation for the READOUT function to preserve maximum capacity, we  
use averaging (i.e. mean pooling) to account for different graph sizes in the database. %

To build one-class classification into the GIN model, we borrow the idea from DeepSVDD \cite{ruff2018deep}. Specifically, we optimize the one-class deep SVDD objective at the output layer of the GIN model as
\begin{align}
	\min_W \quad \frac{1}{N} \sum_{i=1}^N \Vert \text{GIN}(\mcG_i; W)  - \mathbf{c} \Vert^2 + \frac{\lambda}{2} \sum_{l=1}^L \Vert W^l \Vert^2_F
\end{align}
where $W^l$ denotes the parameter of GIN at the $l$-th layer, $W=\{ W^1,...,W^l \}$, and $\mathbf{c}$ is the center of the hypersphere in the representation space that is obtained as the average of all graph representations upon initializing GIN model. Note that the second term corresponds to weight decay of deep models. As mentioned in \cite{ruff2018deep}, deep one-class classification suffers from \textit{feature collapse}, where the trained model maps all input instances to the (constant) $\mathbf{c}$. We employ the regularizations proposed therein to prevent this problem. After model training, the distance to center is used as the outlier score for each graph.

\section{Using Classification Datasets for Outlier Model Evaluation: Issues}
\label{sec:method}

In this section we present in more detail the peculiar ``performance flip'' issue and related observations. Empirically we have observed that the issue is widely existing in many two-stage methods (see Table \ref{tab:all_xnonx}\&\ref{tab:all_xy}) and GNN models (we have evaluated a number of GNN model variants for GLOD in developing new algorithms, although we only present OCGIN.) on lots of datasets. Thus, having a clear understanding of this issue becomes critical for effectively and fairly evaluating detection models and consequently, being able to design better detectors and new models for GLOD.

In the following, we present the peculiar observations in detail (Sec. \ref{ssec:obs}), state our hypothesis on the driving mechanisms behind these observations (Sec. \ref{ssec:hypothesis}), and introduce qualitative and quantitative measures for our empirical analysis (Sec. \ref{ssec:measures}). 
In Sec. \ref{sec:exp} we provide a measurement study using the measures proposed in Sec. \ref{ssec:measures} for propagation based methods to verify our hypothesis, as the underlying mechanism is consistent and easier to analyze. Also the close connection between GNN models and propagation based graph kernels \cite{zhang2018end} strengthens the importance of studying propagation methods. For the other two-stage methods, based on Graph2Vec and FGSD, we provide comprehensive performance evaluation and dicuss its implications. We omit the measurement study, however, as it is harder to analyze their underlying mechanisms. 

\subsection{Peculiar Observations}
\label{ssec:obs}
Graph classification is a widely studied problem with many public datasets available. As graph-level outlier detection is rarely studied with no available dataset, repurposing the graph classification datasets by downsampling one class as outlier can easily provide outlier detection tasks based on real-world samples. For binary classification dataset, there are two ways to down-sample (either down-sample the first class or the other class) to create two variants of outlier detection datasets. In this section, for the first time, we report several unexpected behaviors on various datasets created in this manner, with fixed downsampling rate $0.1$. We split these binary classification datasets into two types: ``X\&Y'' type and ``X\&Non-X'' type. ``X\&Non-X'' refers to datasets with one class representing a category (call it $X$) and the other class representing samples from any other categories other than $X$. ``X\&Y'' type datasets have two classes specifically associated with two different real-world categories $X$ and $Y$ respectively.    

{\bf Setup.~} We conducted GLOD task over 10 datasets (5 ``X\&Y'' and 5 `X\&Non-X'') using a total of 11 GLOD detectors. See Sec.\ref{ssec:setup} for the detailed description, summary statistics of the datasets, and model configurations, and see Table \ref{tab:all_xnonx} and Table \ref{tab:all_xy} for comprehensive results. We summarize peculiar observations across datasets and GLOD detectors. We also give as examples to illustrate our observations on 4 datasets (DD, PROTEINS, NCI1, and IMDB, where IMDB is ``X\&Y'' type and all others are ``X\&Non-X'' type, with performance flip and related issues observed for the first 3 datasets but not for IMDB.), using 3 propagation based detectors (WL+LOF, PK+LOF, and OCGIN) due to space limitation.   



\subsubsection{Peculiar Observation 1: Performance Flip.~} 
The main observation we make in this work is that {\bf for any given outlier detecting models, the performances of detecting outlier appear to depend significantly on \textit{which} class is down-sampled, resulting a large performance gap between two down-sampled variants.}.
Table \ref{tab:flip} shows the ROC-AUC on 4 datasets with their 2 variants at down-sampling rate 0.1 for 3 outlier detecting models. Results for other datasets and models are available at Table \ref{tab:all_xnonx} and Table \ref{tab:all_xy}, with performance flip scenarios marked yellow.  
  
\begin{table}[!t]
    \caption{Average ROC-AUC performance (and standard deviation) of 3 different graph embedding based methods for graph outlier detection using 4 binary graph classification datasets. Each dataset has 2 down-sampled variants, where outliers are created by down-sampling one of two classes (class 0 or class 1) with rate$=0.1$, averaged over 10 different down-samplings. \textbf{Performance flip observed} on DD, PROTEINS, and NCI1 for all 3 models,
    {\bf where ROC-AUC is significantly larger on one variant than the other.}
     ROC-AUC values less than 0.5 are shown \textbf{in bold} as they indicate worse-than-random performance. 
     }\label{tab:flip}
    \centering
    \begin{normalsize}
    \begin{tabular}{lcccc}
    \toprule
     Dataset & Outlier Cls  & OCGIN & WL+LOF & PK+LOF \\
    \midrule
    \multirow{2}{*}{DD} & 0& \BFSERIES 0.327 (0.023) & \BFSERIES  0.186 (0.024) &  \BFSERIES 0.194 (0.027)\\
                        & 1& 0.720 (0.035) &   0.815 (0.020)&   0.824 (0.021)\\
    \midrule
    \multirow{2}{*}{PROTEINS} & 0 &\BFSERIES 0.370 (0.037)& \BFSERIES  0.276 (0.021)& \BFSERIES  0.389 (0.054)\\
                              & 1 & 0.681 (0.028)&   0.664 (0.024)&   0.557 (0.041)\\
    \midrule
    \multirow{2}{*}{NCI1} & 0& 0.643 (0.030)&   0.730 (0.012)&   0.678 (0.019)\\
                          & 1& \BFSERIES 0.467 (0.028)& \BFSERIES  0.349 (0.022)& \BFSERIES  0.366 (0.027)\\
    \midrule\midrule
    \multirow{2}{*}{IMDB} & 0& 0.643 (0.039)&   0.603 (0.038)&   0.624 (0.030)\\
                          & 1& 0.508 (0.049)&   0.651 (0.022)&   0.581 (0.042)\\
    \bottomrule
    \end{tabular}
    \end{normalsize}
\end{table}

\begin{customthm}{1.1}[Performance Gap]
\label{obs:gap} 
A large ROC-AUC gap is widely observed between the two different down-sampled variants of most datasets, consistently across all models. 
\end{customthm}

Strikingly, not only the methods perform well on one variant and poorly on the other, but their performance is simply worse than random (!) in the latter case -- since random ordering would achieve an ROC AUC of 0.5 in expectation.
Perhaps more intriguingly:
\begin{customthm}{1.2}[AUCs sum approximately to 1]
\label{obs:sum1}
The sum of the two ROC-AUC values on the two variants of each dataset is approximately equal to 1, consistently across all models. 
\end{customthm}

To understand this better,
recall the probabilistic interpretation of the ROC-AUC: it is the probability of correctly ranking a random positive instance (i.e. outlier) above a random negative instance (i.e. inlier).
Then these two observations together, which we refer to as the ``performance flip'' issue, suggest a \textit{revised} statement: {\bf the models always consider the graphs from one \textit{fixed} class to be more outlier than those from the other, irrespective of which one is down-sampled.} In other words, it is not that the down-sampled class has impact on the performance, in contrast, the ranking by the models is agnostic to this so-called ``ground-truth'' but rather has a pre-determined bias toward one (fixed) class. Notice that this happens to be class 0 on DD and PROTEINS and class 1 on NCI1 for the three detectors.

\begin{customthm}{1.3}[Performance flip is more severe for ``X\&Non-X'' datasets]\label{obs:gap-degree}
For ``X\&Non-X'' type datasets, the performance flip occurs more often and more severely   (more often having larger performance gap) compared to ``X\&Y'' type datasets.
\end{customthm}
In Table \ref{tab:flip}, the 3 datasets with performance flip are all ``X\&Non-X'' datasets while IMDB is ``X\&Y'' type. Furthermore, full results in Table \ref{tab:all_xnonx} and Table \ref{tab:all_xy} also support that performance flip occurs more often and performance gap is larger in distribution (e.g., at rate 67.3\% vs. 30.6\% for performance gap $\ge 0.2$) on  ``X\&Non-X'' datasets than ``X\&Y'' datasets. 
\begin{customthm}{1.4}[Correlation between performance and class semantics]\label{obs:high-roc}
For propagation based methods, down-sampling ``Non-X'' class as outlier in``X\&Non-X'' dataset always achieves high performance. However it is not always true for other methods.
\end{customthm}
By definition ``X'' class refers to a category of instances that likely exhibit characteristic patterns of the class, while the ``Non-X'' class contains many patterns out of the ``X'' class. By aligning the performance result (Table \ref{tab:all_xnonx}) with class semantics shown in Table \ref{tab:data-semantics}, we find clearly that \textit{all} propagation based methods achieve high performance for down-sampling ``Non-X''. This supports our following analysis over propagation based methods, where we find that these methods have larger sparsification rate for more diverse classes.

We conclude with four remarks. First, note that although the ranking of models by performance differs from dataset to dataset, the performance-flip and AUCs-sum-approximately-to-1 behaviors are consistent across models. 
Second, the issue does not appear to be universal as it does not arise on IMDB (cf. Table \ref{tab:flip}), where all models are better than random on both variants. 
Third, when performance flip is observed, which version of the downsample achieving high performance depends on both the embedding method as well as the downstream outlier detector, that is the performance gap can be reversed when using different type of graph embedding. Last, learning-based end-to-end model has the ability of capturing majority class distribution to a certain degree with performance flip not occurring for all ``X\&Y'' type datasets, which points out a potentially promising direction to overcome performance flip.  



\subsubsection{Peculiar Observation 2: Invariance to Down-sampling Rate.~} 
\begin{figure}[h!]
   \begin{tabular}{ccc|c}
    \hspace{-0.1in}\subfloat[DD]{\includegraphics[width=0.25\textwidth]{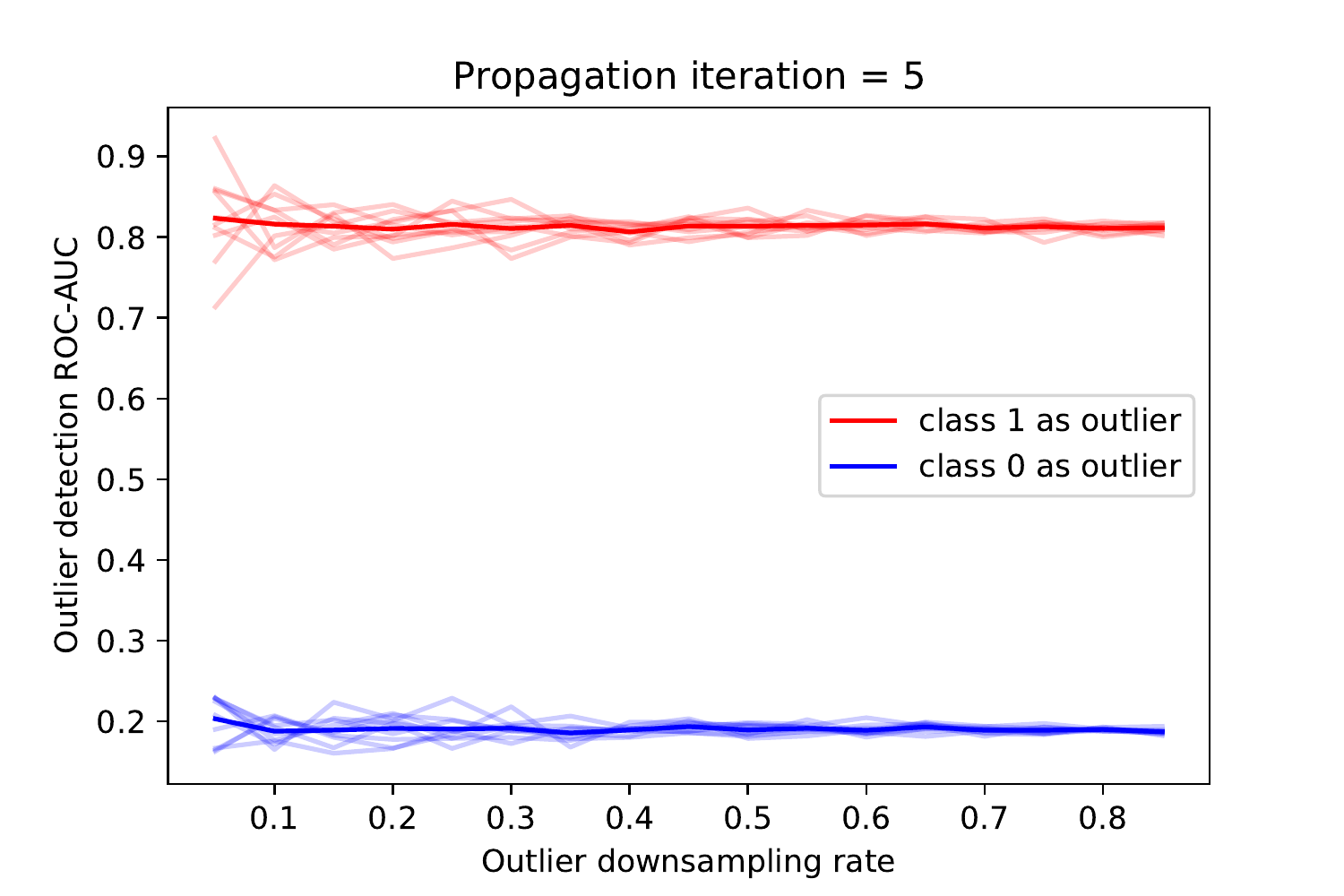}} &
    \hspace{-0.125in}\subfloat[PROTEINS]{\includegraphics[width=0.25\textwidth]{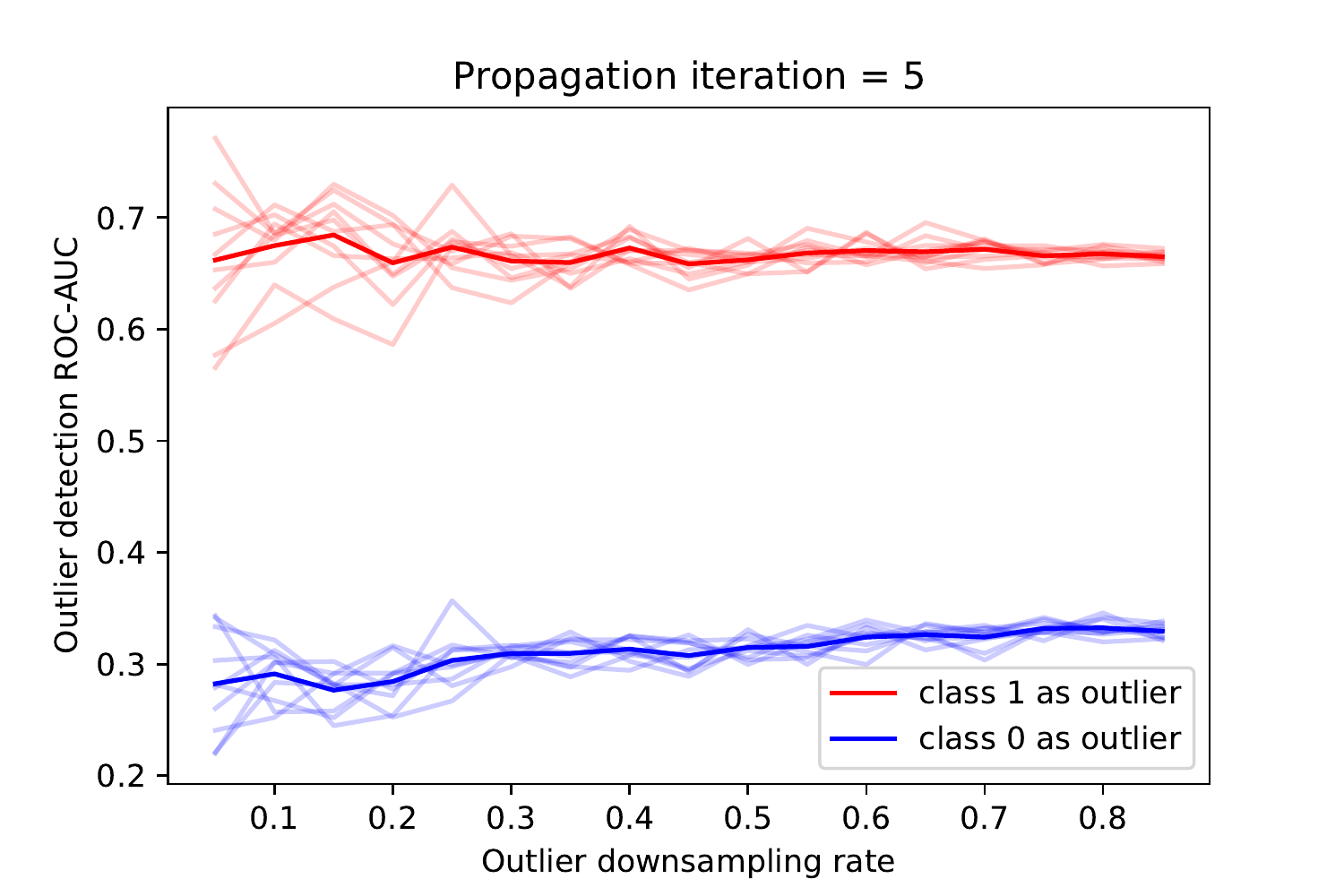}}&
    \hspace{-0.125in}\subfloat[NCI1]{\includegraphics[width=0.25\textwidth]{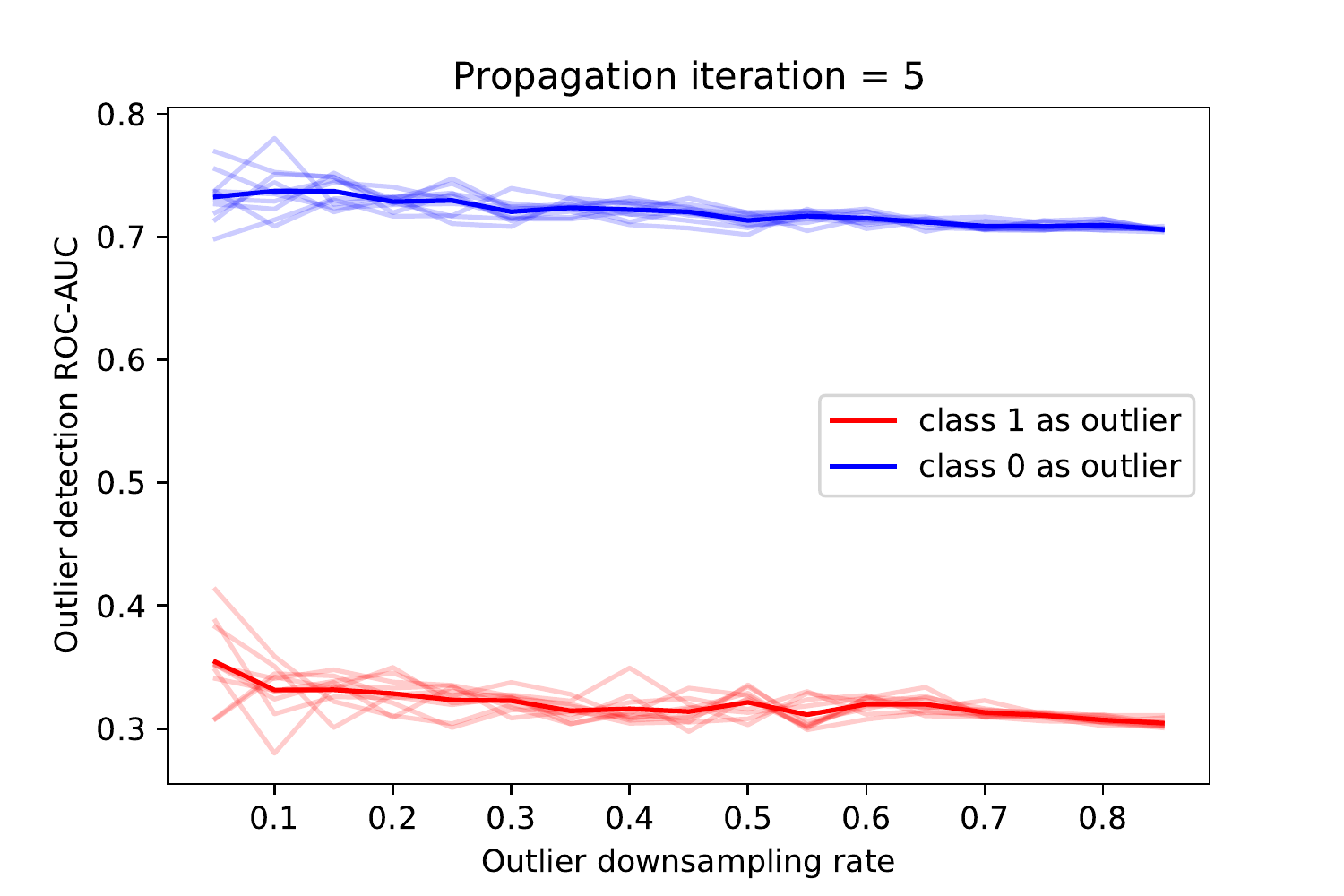}}&
    \hspace{-0.00in}\subfloat[IMDB]{\includegraphics[width=0.25\textwidth]{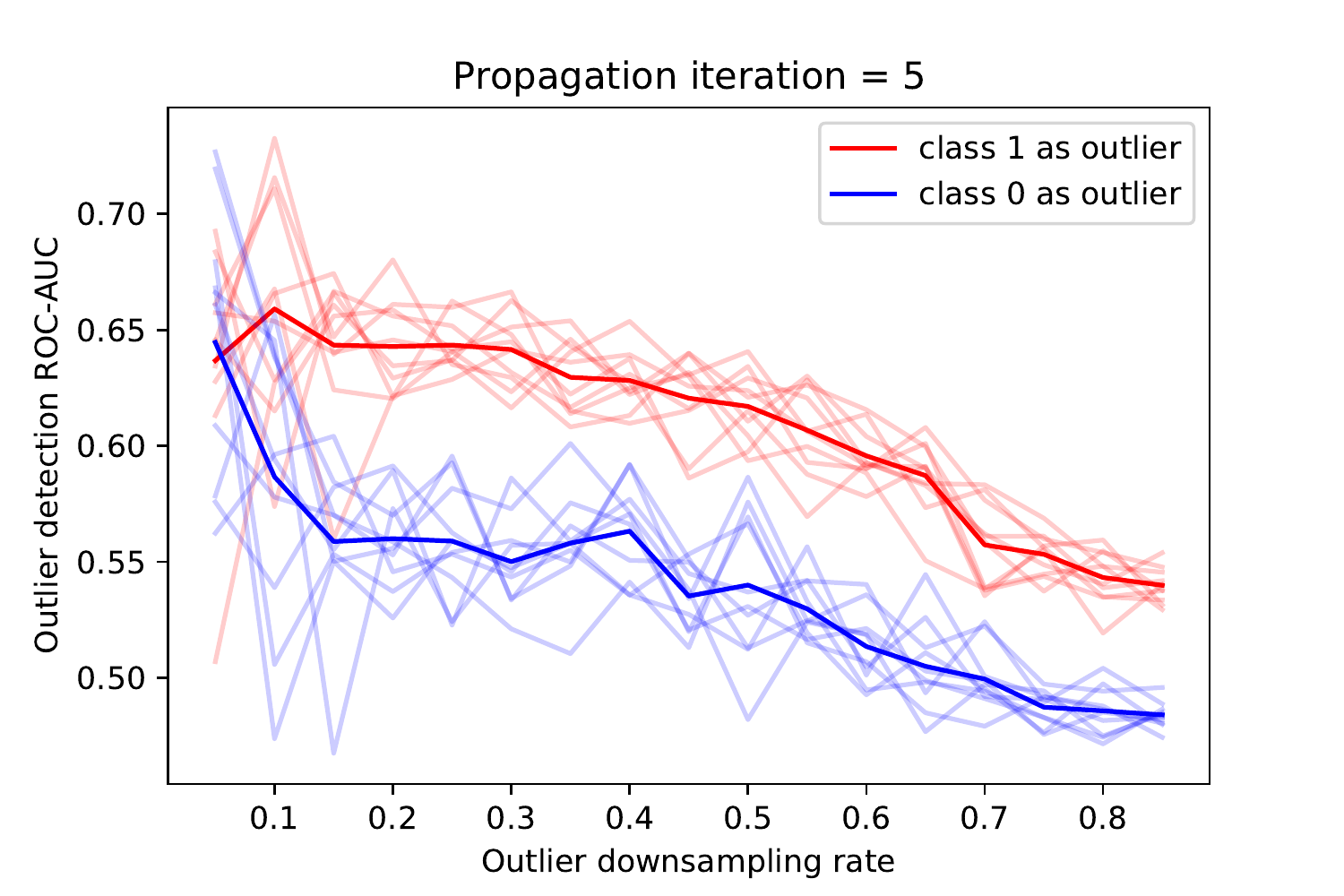}}\\
    \end{tabular}
    \caption{\textbf{Performance is invariant to downsampling rate} on DD, PROTEINS, and NCI1 for WL+LOF. Similar behavior is observed for methods and datasets when performance flip is occur.}\label{fig:downsampling}
     \vspace{-.1in}
\end{figure}

When down-sampling one class as outlier with a certain down-sampling rate, we would conjecture that a lower rate would make the outlier detection task easier as the density of outliers becomes lower. 
Fig. \ref{fig:downsampling} shows the detection ROC-AUC of WL+LOF (with $L=5$ iterations) for various down-sampling rates, from 0.05 to 0.85, on 
both variants of all datasets.
The conjecture appears to hold only for IMDB -- on which performance flip is not observed. In contrast,   
the performance is strikingly flat on DD, PROTEINS, and NCI1. Similar results hold for PK+LOF and OCGIN on these three datasets (See Sec. \ref{sec:exp}). More broadly, the observation holds for all methods and datasets when performance flip is observed (see additional result in our project webpage \projurl). 

\begin{customthm}{2}[Invariance to down-sampling rate]
\label{obs:flat}
 Performance flip issue is not an artifact of the down-sampling rate, in fact, ROC-AUC appears to be invariant to the rate.
\end{customthm}

This observation is in agreement with Observation \ref{obs:sum1} (AUCs-sum-approximately-to-1).
The probabilistic interpretation of ROC-AUC is regarding \textit{any two} random positive-negative instances, irrespective of the total number of instances from those groups.


\subsubsection{Peculiar Observation 3: Growing Performance Gap with Propagation (propagation based methods only).~} 

\begin{figure}[h!]
    \centering
    \begin{tabular}{ccc|c}
        \hspace{-0.1in} \subfloat[DD]{\includegraphics[width=0.25\textwidth]{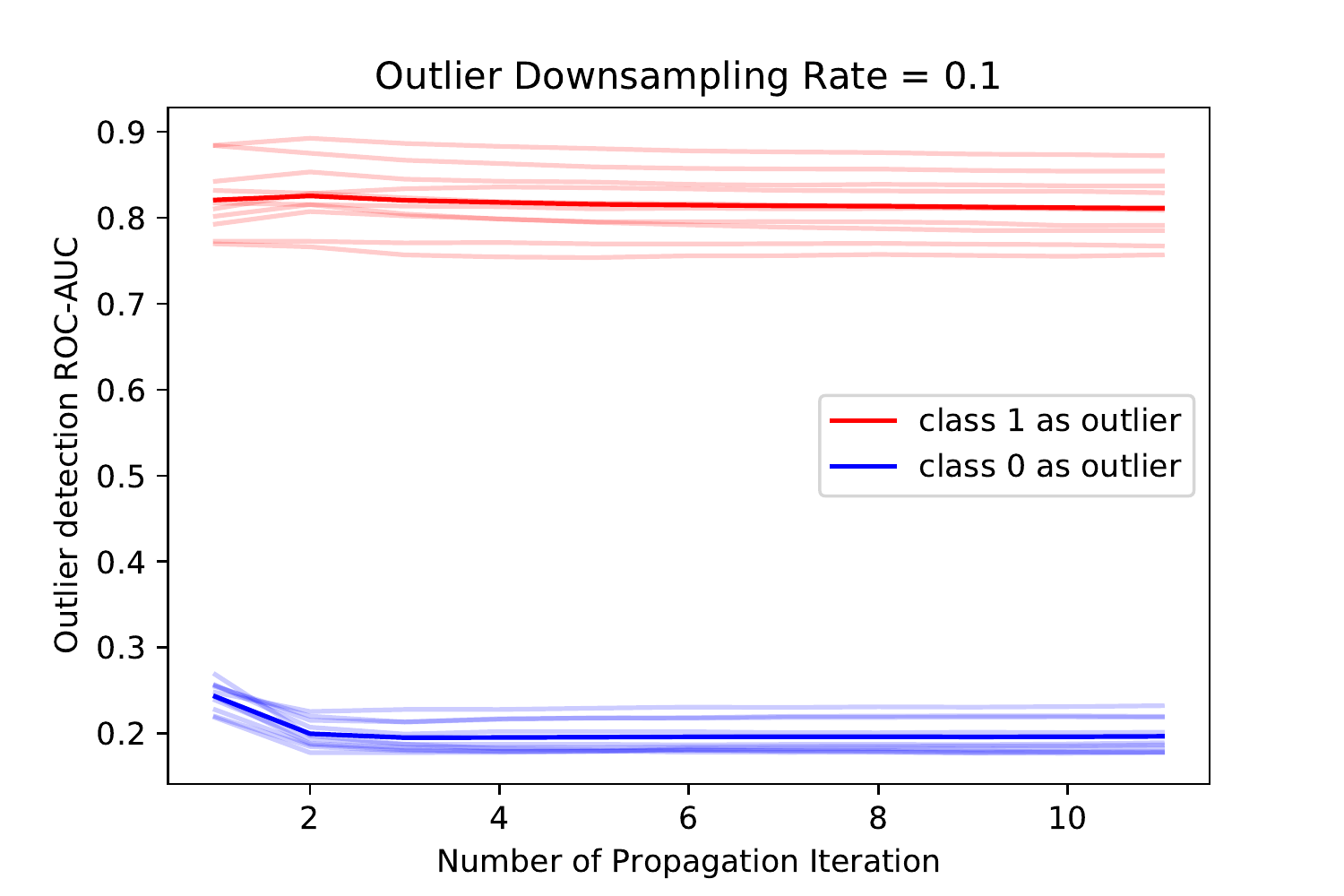}} &
         \hspace{-0.175in}\subfloat[PROTEINS]{\includegraphics[width=0.25\textwidth]{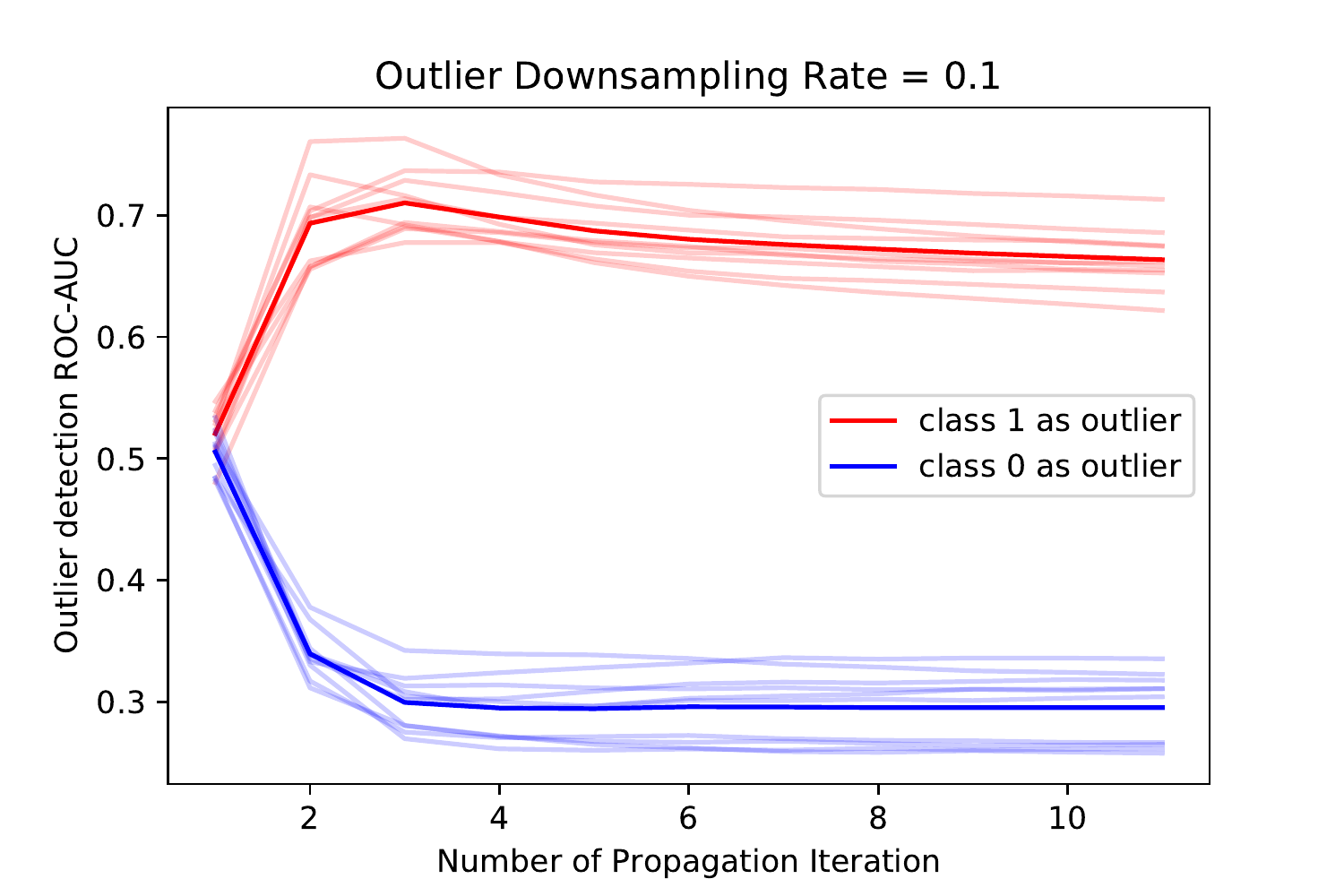}}&
        \hspace{-0.175in} \subfloat[NCI1]{\includegraphics[width=0.25\textwidth]{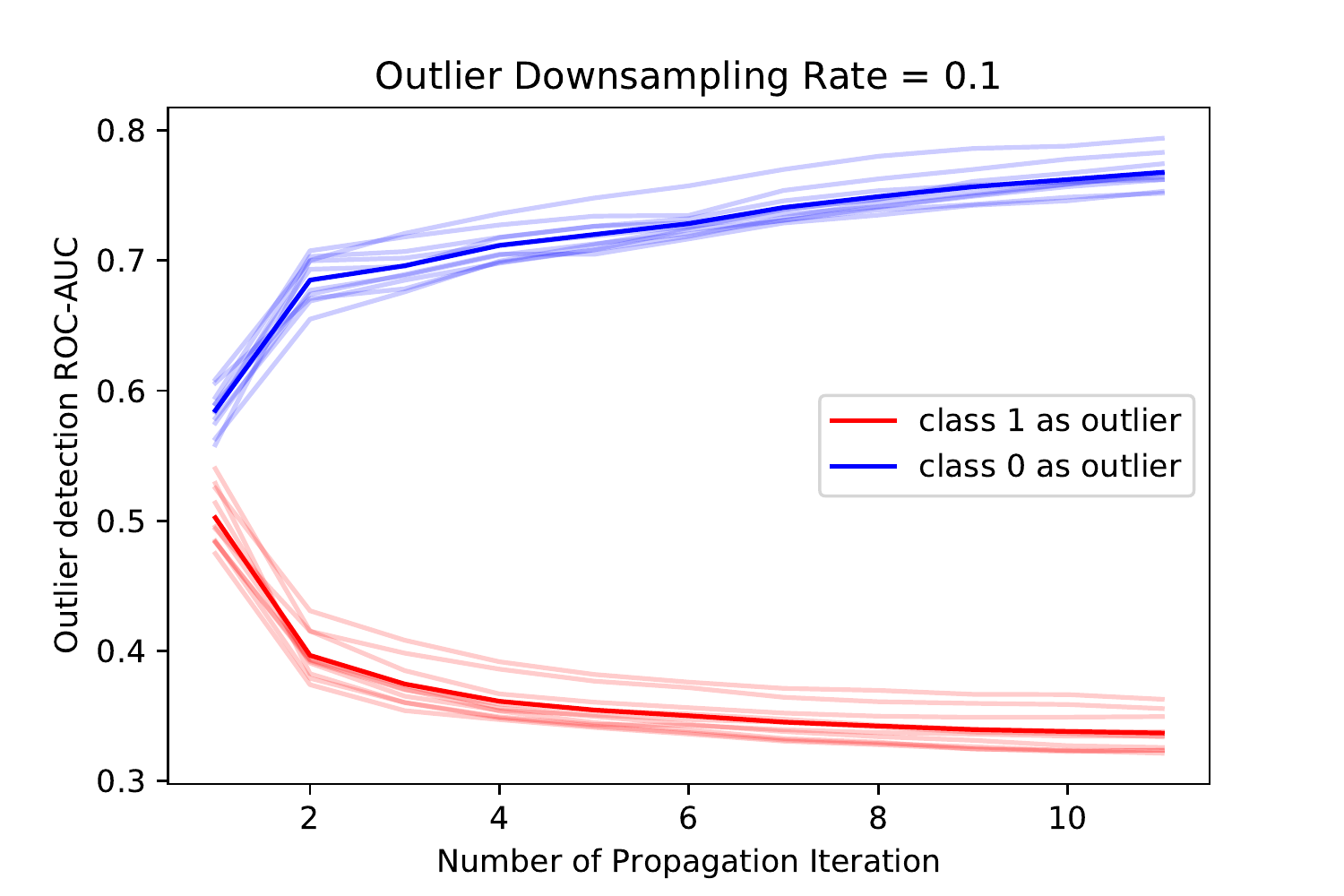}}&
        \hspace{-0.00in}\subfloat[IMDB]{\includegraphics[width=0.25\textwidth]{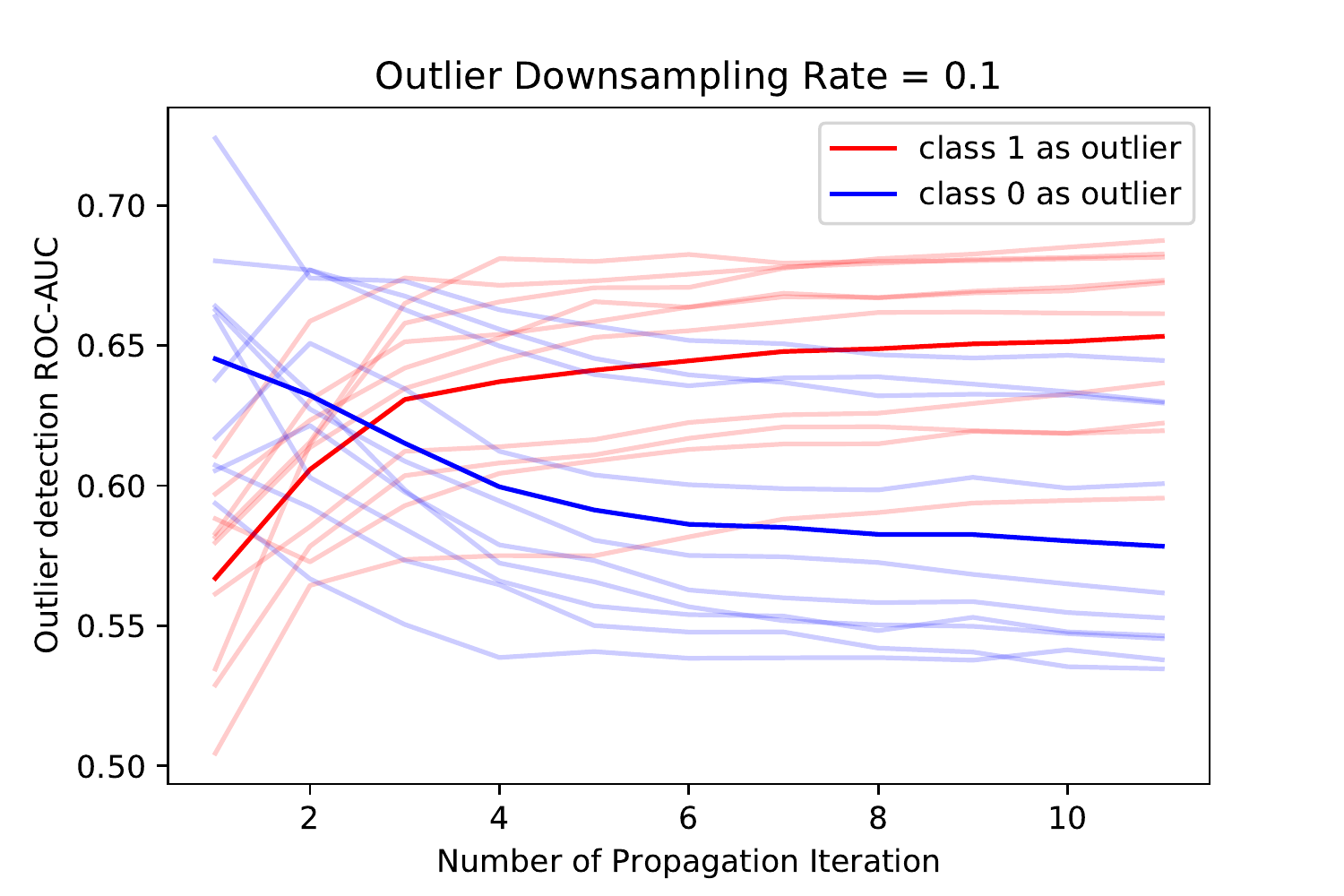}}\\
        \end{tabular}
       \vspace{.05in}
    \caption{\textbf{Performance gap between two variants tends to grow with increasing number of propagations} (i.e. iterations) of WL (subsequently paired with LOF), significantly on DD, PROTEINS, and NCI1. Similar behavior is observed for other propagation based methods and datasets.}\label{fig:propagation}
\end{figure}

Observation \ref{obs:flat} is mainly related to a property of the dataset generation. On the other hand, a key property of the outlier models we employ in this work is the number of iterations (for WL and PK) or the number of layers (for GIN), earlier denoted with $L$ (See Sec. \ref{ssec:models}), both of which correspond to propagations over the graph. Here we look at how the performance behaves under varying $L$. 
Fig. \ref{fig:propagation} shows the performance of WL+LOF on all datasets for two variants for $L$ increased from 1 through 11. Results are qualitatively similar for PK+LOF, however OCGIN behaves differently (See Sec. \ref{sec:exp}).

\begin{customthm}{3}[Growing gap with propagation]
\label{obs:propgrow}
 The difference in ROC-AUC performances on two different down-sampled variants of a classification dataset tends to grow with increasing number of graph propagations for WL- and PK-based outlier models. For OCGIN there exists no obvious growth. 
\end{customthm} 

The growth is significant particularly on DD, PROTEINS, and NCI1 for which the performance flip occurs.
It is interesting that this behavior is mainly associated with two-stage propagation models and not with our end-to-end model. 
We can reason about the two-stage models based on the sparsification property that both exhibit. (In contrast, OCGIN is optimized where it is harder to reason about how the learning of its many parameters affects performance.)
Recall that sparsification, as discussed in Sec. \ref{sec:dominance}, implies that the kernel distance between two graphs tends to increase with number of WL and PK iterations. Combined with Observation \ref{obs:high-roc} we hypothesize that sparsification issue with increasing propagations is significant more noticeable for ``Non-X'' class (one with diverse patterns) and down-sampling this class as outlier results in an easier task where outliers are sparse and dispersed in the embedding space.


\subsection{Hypothesis on Driving Mechanisms}
\label{ssec:hypothesis}


Next we aim to build an understanding  of 
the leading factors behind the unusual observations we have presented. As the underlying mechanism of propagation based methods is consistent and easier to analyze, we present our hypothesis and understanding based on propagation based methods. However the main argument of two factors (density disparity and overlapping support of graph embedding space) holds true for other non-propagation based methods as well.
To that end, we focus on investigating pairwise similarities -- either produced by the graph kernel or the dot product between graph embeddings -- normalized in range $[0, 1]$. We consider all-pairs similarities among all graphs 
before down-sampling. These convey important information about the density of and the distance between the graphs across classes in the graph embedding space, which many downstream outlier detectors rely on, including the two -- LOF and OCSVM -- we have used.


\begin{figure}
	\centering
	\includegraphics[width=\textwidth]{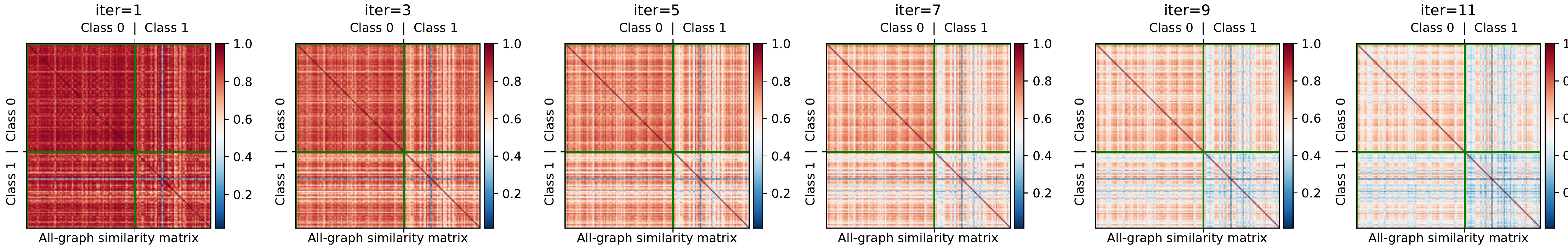}
	\vspace{0.001in}
	\caption{Pairwise similarities among all graphs in DD dataset (graphs grouped by class) based on WL subtree kernel over increasing iterations (left to right).} \label{fig:similarity_matrix}
	\vspace{0.1in}
\end{figure}

Fig. \ref{fig:similarity_matrix} shows the pairwise similarities among all graphs as well as how those change with increasing number of graph propagations (i.e. iterations) based on the WL subtree kernel on DD dataset where 0 represents enzyme and 1 represents non-enzyme. (Similar plots for PK and OCGIN can (will) be found in \projurl \footnote{Our study involves an in-depth study of three different graph outlier detection models. Due to limited space for figures, we include additional or corresponding figures in the Appendix section of our arXiv submission  \cite{zhao2020using}.}) The pairwise similarities are block-wise grouped based on the true class label. We emphasize two factors:

\bit 
\item[1.] \textit{Diagonal, block-wise (intra-class) similarities:~} 
Clearly sparsification arises for \textit{both} classes, where the graphs within the same class become more and more dissimilar to one another as the number of iterations increases.
What is even more important to note is that \textit{the speed at which sparsification occurs is different for the two classes}. 
As shown in Fig. \ref{fig:similarity_matrix} for DD, distribution of graphs in class 1 sparsifies much faster than that for class 0, leading to a {\em disparity between class-level densities}.
Although less apparent from the first subfigure, density disparity exists even after a single iteration. The initial disparity difference aligns with the class semantic meaning where class 0 represents enzyme and is more compact (denser) than class 1. Increasing number of iterations only amplifies this disparity.

\item[2.] \textit{Off-diagonal (inter-class) similarities:~} Pairwise similarities among graphs within the sparser class (in Fig. \ref{fig:similarity_matrix}, class 1 for DD) are {\em lower} than inter-class similarities on average.
Put differently, graphs within the sparser class tend to look more similar to graphs in the other class than among themselves.
This suggests that the class-level distributions of graphs in the embedding space have {\em overlapping support}.
\eit

Based on the above, we conjecture the following conditions as the leading factors behind our peculiar ``performance flip'' and related observations:
\bit
	\item \textbf{(Growing) Density Disparity.} 
	Density of outliers in the feature space is often inversely correlated with the difficulty of identifying them.
	That is, the more spread out and apart from inliers are the outliers, the easier the detection task gets.
	This suggests that designating as outliers the graphs from the class with sparser graph embeddings (or the class that sparsifies faster) would induce a relatively easier task, as compared to graphs from the other class with higher embedding density. What is more, graph propagation would contribute to a growing disparity between task difficulties, as density disparity grows with more propagation. For propagation based graph embedding methods the class density over embedding space aligns with the diversity of patterns in the original graph space, but this does not always hold for other graph embedding methods.
	

	\item \textbf{Overlapping Support.} 
	Density disparity alone is not a sufficient condition to explain the performance disparity (or flip) issue.
	Two sets of graph embeddings that are fully separable in the feature space would both induce an `easy' task, no matter how different their within-set (i.e. class) densities are. {Unsupervised}  graph representation learning methods, however, do not necessarily have the ability (at least explicitly) to embed graphs from the same class closeby while maintaining inter-class separability. On the contrary, they tend to generate mixed embeddings that have a common/overlapping support among classes.

\eit

We conclude by forming the following hypothesis on the driving mechanisms behind ``performance flip'' and related observations.
 Provided graph embedding methods employed for outlier detection induce both density disparity as well as overlapping support,
 one down-sampling scenario creates a dense inlier distribution surrounded with dispersed outliers (`easy' task), whereas the other creates a sparse inlier distribution that has overlapping support with a small set of outliers with relatively higher density (`hard' task). Since most models assume the former scenario in their formalism, they `do well' on the respective `easy' task, and poorly on the other (Observation \ref{obs:gap}). 
More broadly, therefore, the performance flip issue stems from the (mis)alignment of embedding (i.e. inlier/outlier) distributions with the underlying assumptions of the detection models.
As the embedding space is produced by graph embedding methods, which class has denser embeddings relies on the specific graph embedding method and may not align with semantic meaning of class labels. However when the original dataset has one class with clearly more diverse patterns (like ``Non-X'' class in ``X\&Non-X'' datasets), the propagation based methods in general produce sparser embedding space for the semantically-more-diverse class (Observation \ref{obs:high-roc} and Observation \ref{obs:gap-degree}). 
When performance flip happens, perhaps more disturbingly, the models tend to always consider the graphs from the sparser class to be more outlying than the graphs from the other, denser class -- no matter {\em which} one is down-sampled and at what rate (Observation \ref{obs:flat}). That is, their probability of ranking a sparser-class instance above a denser-class instance remains almost the same, which leads to the observed AUCs-sum-approximately-to-1 behavior (Observation \ref{obs:sum1}) and consequently, the worse-than-random performance when the denser class is down-sampled. The issue is exacerbated with more graph propagation as it leads to a growing disparity of densities and respective task difficulties  (Observation \ref{obs:propgrow}).




\subsection{Measures for Analysis}
\label{ssec:measures}
In the previous section we pointed out overlapping support and (growing) density disparity to be two key factors leading to the observed unusual behaviors. 
Here we introduce concrete measures to quantify these two factors.
\begin{enumerate}
	\item \textbf{Qualitative visualization of overlapping support and growing density disparity:} 
	Pairwise similarities do not directly reflect how points (i.e. graphs) are distributed in the representation space. 
	We use multi-dimensional scaling (MDS) to map the graph embeddings into 2-dimensions (MDS mapping aims to preserve the relative pairwise similarities) wherein points from different classes are colored differently. 
 Overlapping support can be validated visually from the mixing of two colors. On the other hand, the growing density disparity
 can be supported by the varying spread of points across classes and the rate at which this spread changes for each class across MDS visualizations corresponding to increasing propagation by a detector.
\end{enumerate}
	
	We remark that the MDS visualizations provide only qualitative evidence, as the mixing of colors and the varying spread of points could also be attributed to MDS error, i.e. could simply be artifacts; due to the space in which the pairwise similarities are to be preserved is constrained to only 2 dimensions for visualization purposes.
	Therefore, we also analyze quantitative measures of these factors described as follows.
	
\begin{enumerate}[resume]
	\item \textbf{Quantitative measure of density disparity:} 
	The distance between any two graphs from the sparse class would be larger. Therefore, we use the so-called  	
	\textit{NN-Radius} to quantify the degree of density.
	\begin{definition} 
		\textit{NN-Radius is the distance (or $1$$-$similarity as normalized in range $[0, 1]$) to the $k$-th\footnote{In the paper we report results for $k=20$ and note that the take-aways are not sensitive to this choice.} nearest neighbor (NN) of a graph in the embedding space.}
	\end{definition}
	A larger radius corresponds to lower local density. The distribution of the NN-radii of the graphs from each class measures the density of the class. A more left-shifted distribution would imply a denser class. 

	\item \textbf{Quantitative measure of overlapping support:} 
	In the absence of overlap, assuming graphs from different classes are linearly well-separated in the embedding space (forming two disjoint clusters), we would expect the nearest neighbors of each graph to be from the same class. Therefore, we use the so-called {\em NN-Disagreement\%} as the degree of overlap.
	\begin{definition} 
			\textit{NN-Disagreement\% is the percentage of graphs from the opposite class within the NN-Radius of a graph.}
	\end{definition}	
	We then study the distribution of NN-impurities of the graphs from each class.
	A left- or right-shifted distribution would respectively show whether a class is more likely to be surrounded by its own members (i.e. well-clustered, dense) or not (i.e. dispersed, sparse).
\end{enumerate}

 \subsection{A deeper analysis on embedding sparsity issue}
 \label{ssec:simulate}

In this section we perform various simulations to better understand the disparity in sparsification rates between two classes, that is how one sparsifies faster than the other (See e.g. Fig. \ref{fig:similarity_matrix}).
We use WL subtree kernel for the simulations as these do not require any parameter training and hence are easier to analyze. 
The simulations are designed to show: (1) Even a small difference between two graphs can be amplified by propagation, leading to sparsification; and (2) Various factors, including differences in label distribution and topology, contributes to differences in 
the rate of sparsification between two classes.

We consider simulations on $k$-regular graphs where each node has exactly $k$ number of neighbors, i.e. the same degree.
(Unlabeled) $k$-regular graphs are among the class of graphs on which 1-dimensional WL (1-WL) isomorphism test \cite{weisfeiler1968reduction} fails to reject non-isomorphism. That is, 1-WL test cannot distinguish two structurally non-isomorphic 
\kreg~ graphs.
In practice, node/edge labels or attributes may help diminish such difficulty by breaking the symmetry \cite{2020arXiv200900142L}. This is exactly the route we take to induce asymmetry in a controlled fashion (from low to high). Specifically, we design two cases:

\vspace{0.1in}
\bit
\item {\bf Case 1:} We start with two identical $(k,n)$ regular graphs ($n$ nodes, all with degree=$k$), where all nodes are labeled $A$. 
Then, we flip the labels of $m$ randomly chosen  nodes in each graph to $B$.
As such, the graphs are structurally isomorphic, whereas $B$-labeled nodes are ensured to 
induce assymetry (and hence non-isomorphism) between the graphs.

\vspace{0.05in}
\item {\bf Case 2:} We start with two identical $(k,n)$ regular graphs, where the nodes are labeled with two different labels, $A$ or $B$. To induce non-isomorphism, we perform degree-preserving rewiring between $r$ randomly chosen edge pairs in one of the graphs.
\eit 
\vspace{0.1in}

\begin{figure}[h!]
\centering
	\begin{tabular}{cc}
	\includegraphics[width=0.35\textwidth]{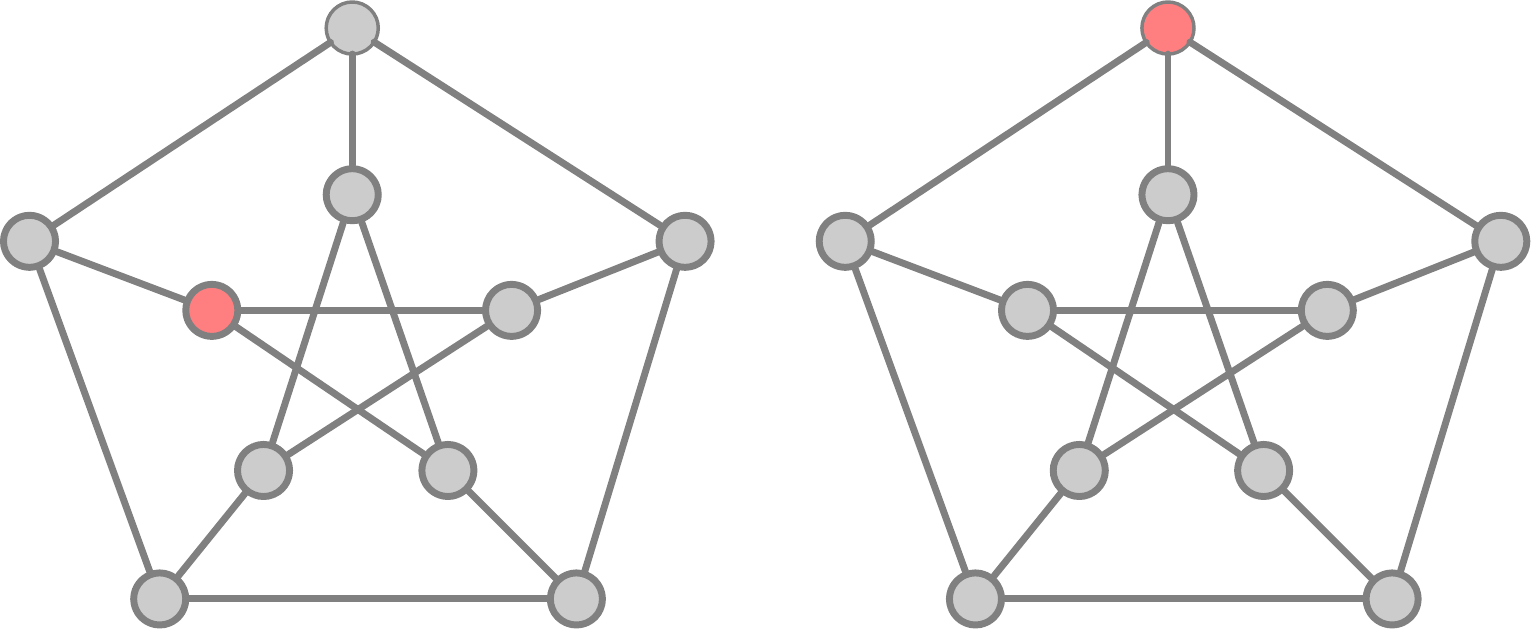} &
	\includegraphics[width=0.35\textwidth]{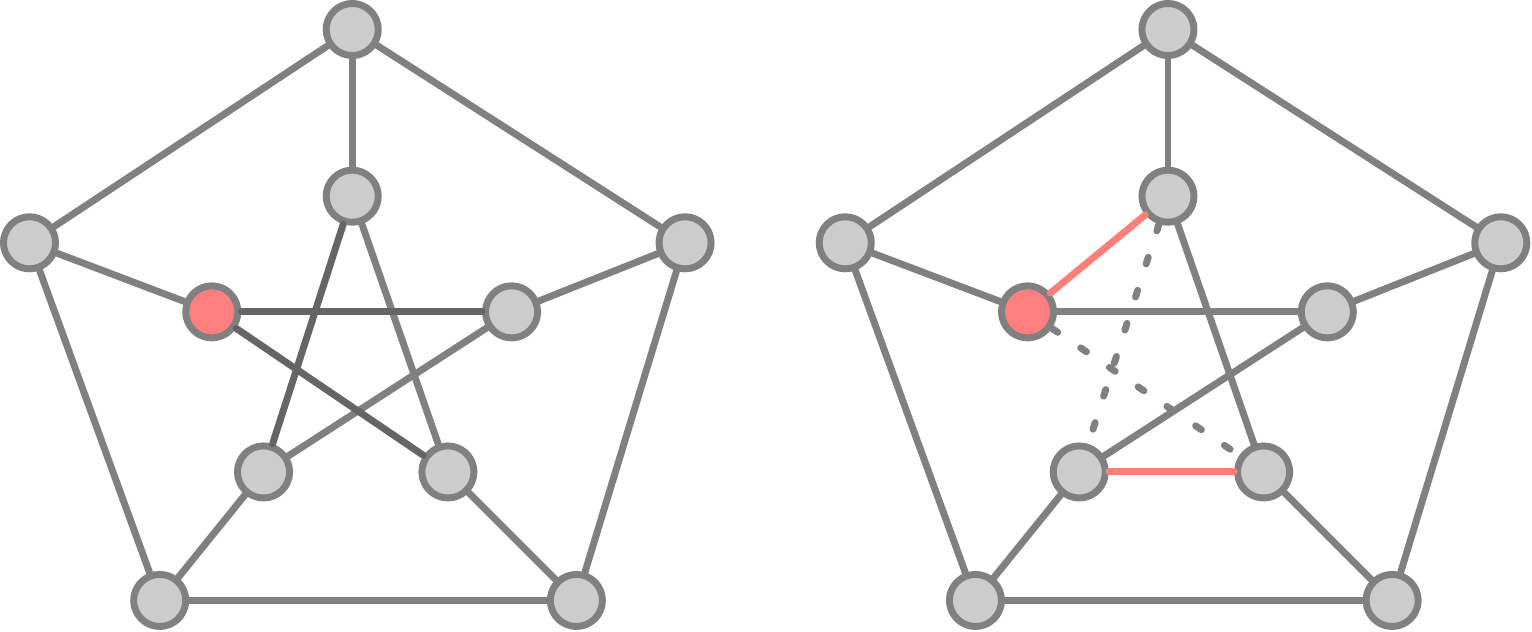}\\
	(a) non-isomorphism by node labeling & (b) non-isomorphism by edge rewiring \\
	\end{tabular}
\caption{Non-isomorphism is induced (a)  between two unlabeled Petersen graphs by flipping node labels assymmetrically, and 
(b) between two labeled Petersen graphs by degree-preserving edge rewiring.} \label{fig:petersen}
\end{figure}

Our goal is to start from a place where WL distance is zero (i.e. identical graphs) and to study the effect of increasing $m$ or $r$ on the distance growth.
Fig. \ref{fig:petersen}(a) illustrates Case 1 on two so-called Petersen graphs ($k=3$, $n=10$) where $m=1$ node's label in each graph  is flipped from gray (label $A$) to red (label $B$), inducing non-isomorphism.
Case 2 is illustrated in (b) where we rewire $r=1$ edge pair in one graph (right) to create a graph that is non-isomorphic to the other (left) graph.

\vspace{.05in}
\vspace{.05in}

Without the perturbations (label flipping in Case 1, and rewiring in Case 2), WL distance  between the two graphs would be equal to zero in both cases, irrespective of number of iterations.
Just a small change, as shown in Fig. \ref{fig:petersen}, ``jump-starts'' WL where the distance between the graphs starts growing with increasing WL iterations. 

\begin{figure}[h!]
\centering
	\begin{tabular}{cp{0.05cm}c}
		\includegraphics[width=0.425\textwidth]{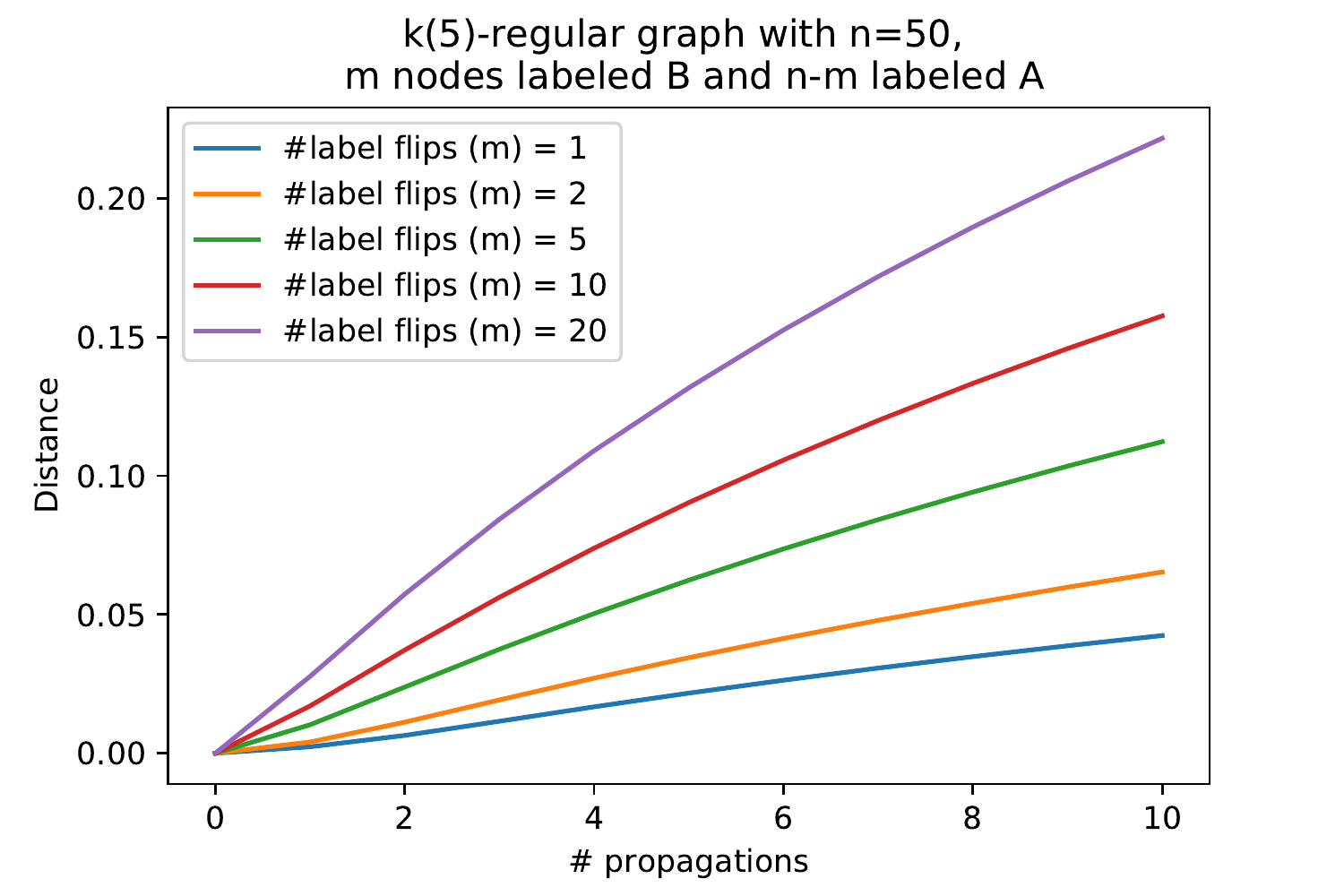} &&
		\includegraphics[width=0.425\textwidth]{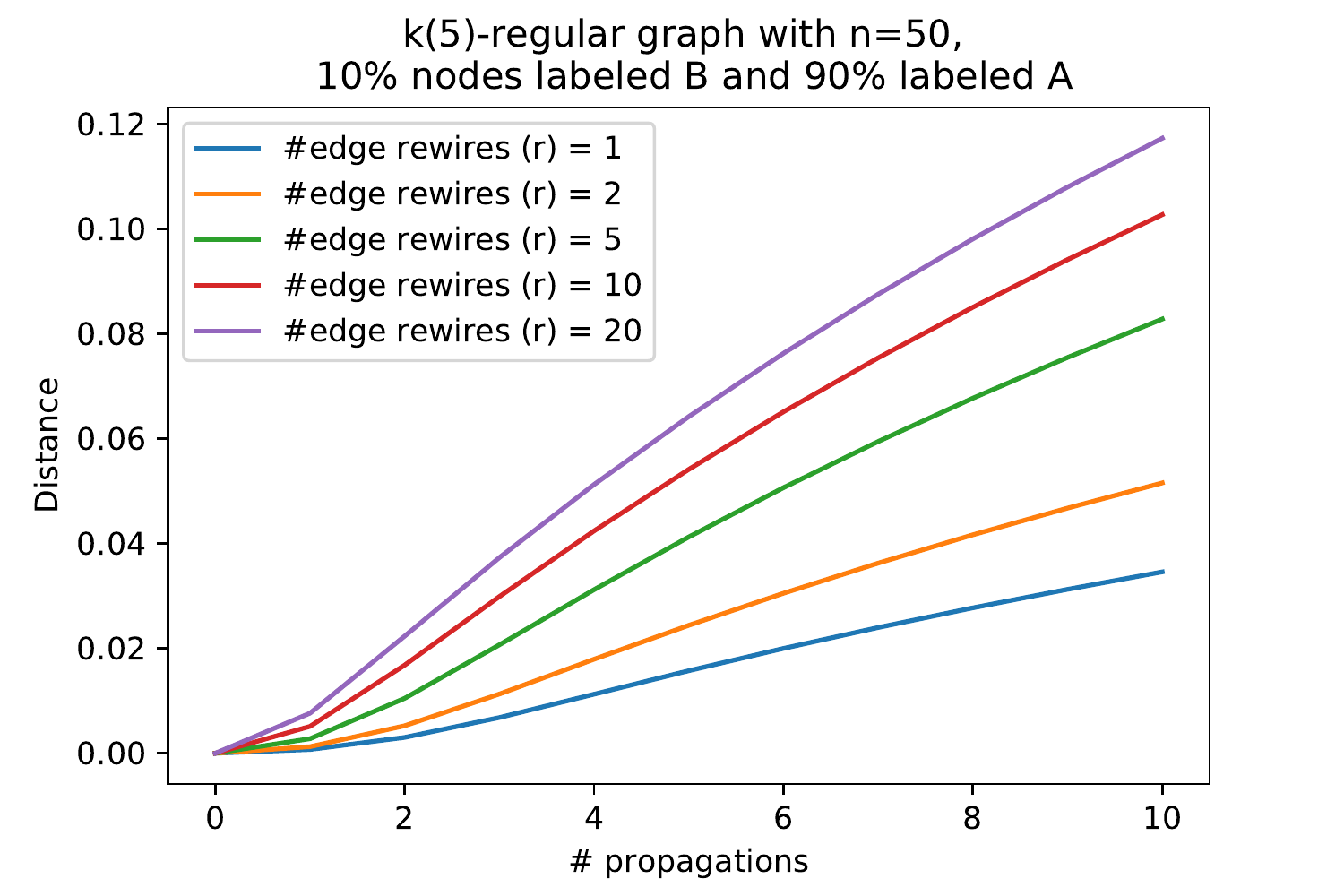}\\
	\end{tabular}
	\vspace{.05in}
	\caption{Distance between two \kreg~graphs ($k=5$, $n=50$) as a function of (left, Case 1) number of node labels flipped, and (right, Case 2) number of edge pairs rewired. Each curve is averaged over 100 rounds to remove randomness.} \label{fig:changingm}
	\vspace{.05in}
\end{figure}

In Fig. \ref{fig:changingm} (a) and (b) we show how graph distance changes with increasing WL iterations as a function of number of node labels flipped in Case 1 and number of edge pairs rewired in Case 2, respectively.
Notice that even after a relatively small perturbation on otherwise non-distinguishable graphs, their distance grows over iterations. The bigger the amount of perturbation (i.e. the difference between two graphs), the larger the distances as reflected by WL, and also the larger the growth rate (note the increasing gap between curves). 


\begin{figure}[h!]
\centering
	\begin{tabular}{cp{0.05cm}c}
		\includegraphics[width=0.425\textwidth]{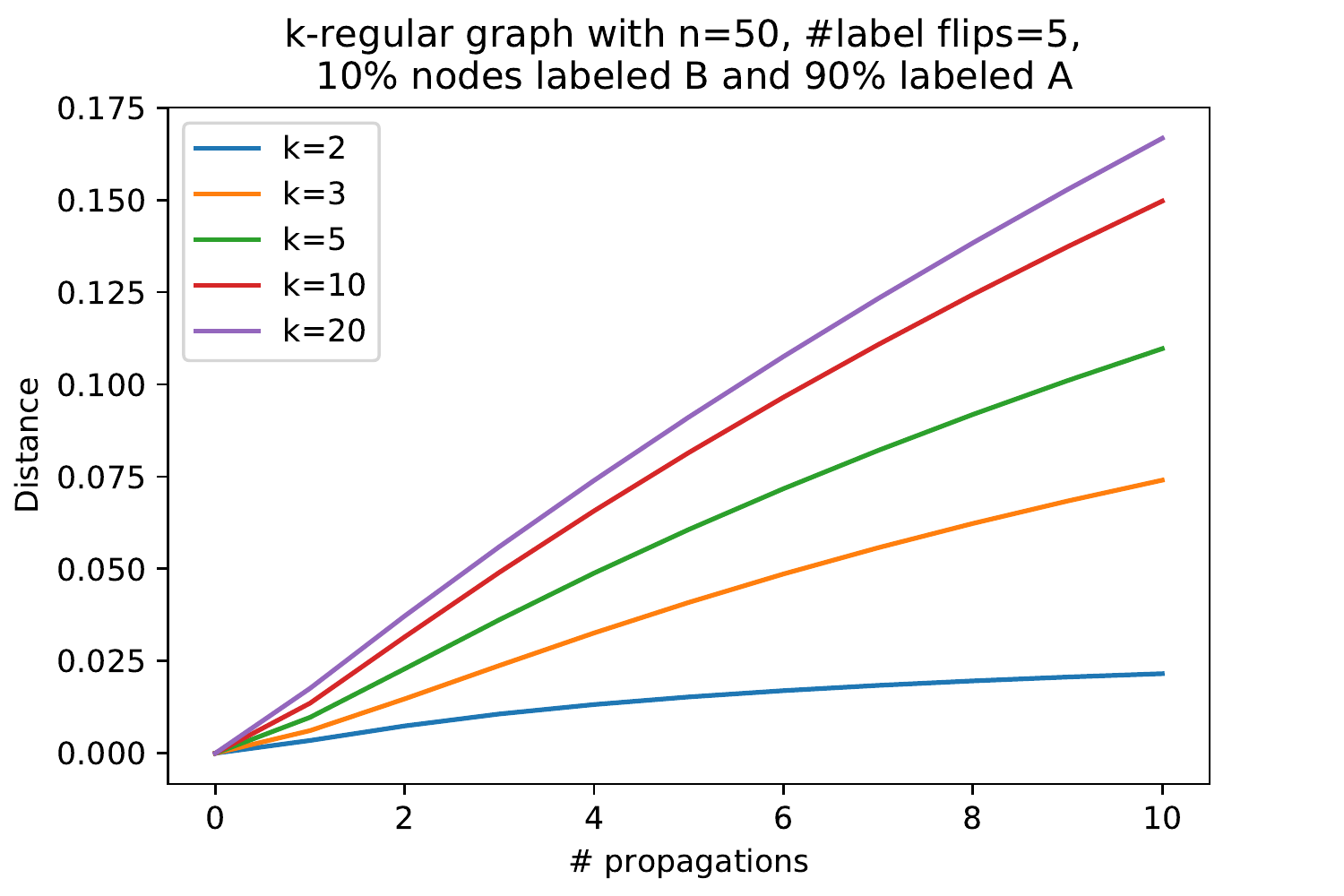} &&
		\includegraphics[width=0.425\textwidth]{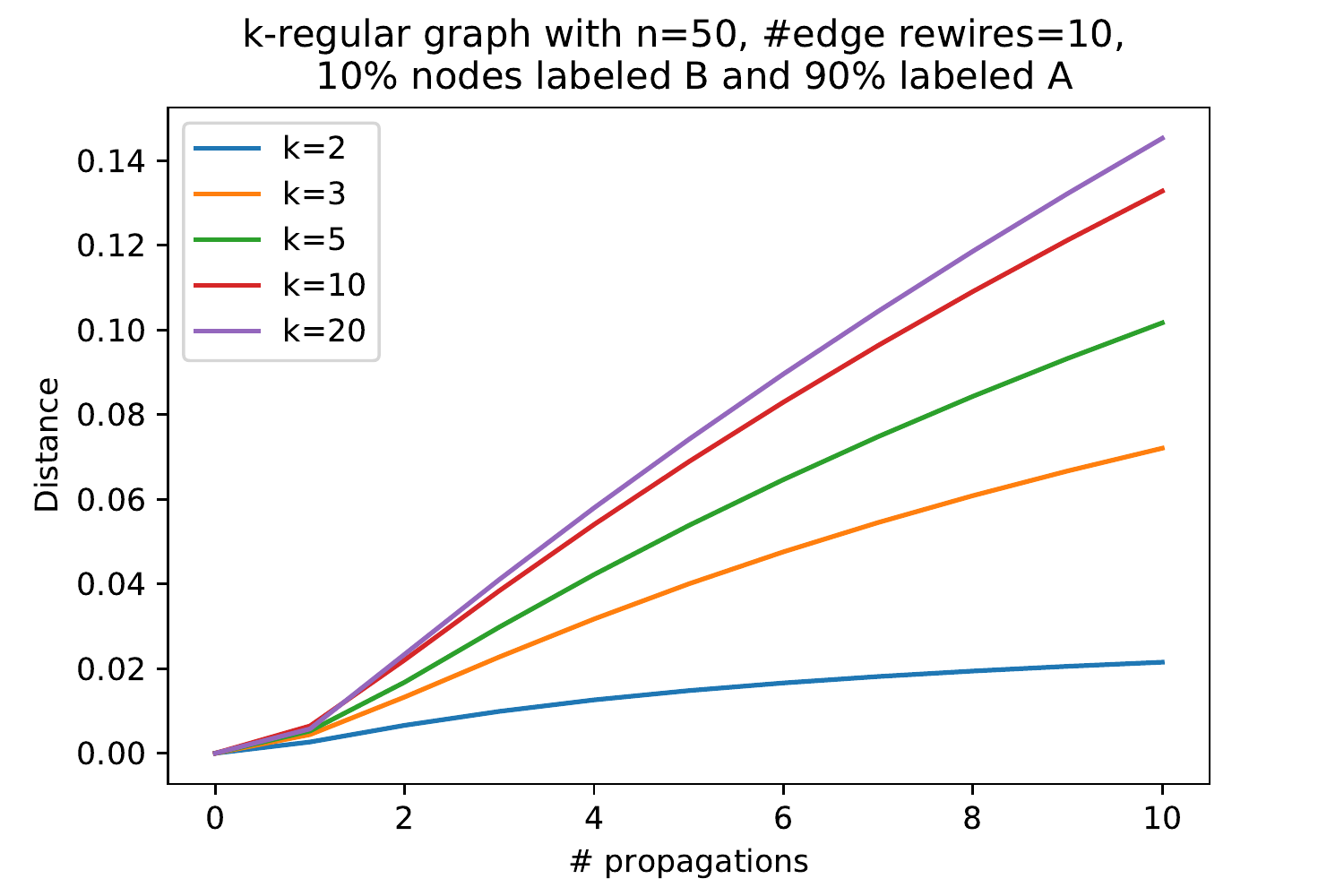}\\
	\end{tabular}
	\vspace{.05in}
	\caption{Distance between two \kreg~graphs ($n=50$) as a function of $k$ for (left, Case 1) when $m=5$ node labels are flipped, and for (right, Case 2) when $r=10$ edge pairs are rewired. Each curve is averaged over 100 rounds to remove randomness.} \label{fig:changingk}
	\vspace{.05in}
\end{figure}

The distance growth is also a consequence of graph topology.
Fig. \ref{fig:changingk}(a) shows graph distance over WL iterations on \kreg~graphs with varying $k$, when number of label flips is fixed to $m=5$ in Case 1.
Similarly, Fig. \ref{fig:changingk}(b) shows the same when the number of edge rewirings is fixed to $r=10$ in Case 2.
In both cases the graph distances are larger for increasing $k$ across iterations.
These show that the effect on distance of the same (even a small) amount of perturbation on two graphs varies depending on the topology.

All in all, sparsification is evident from our simulations -- i.e. distances grow over iterations, or graph propagations. The growth rate of sparsification is sensitive to various factors (difference in label distribution, difference in topology, etc.). It is unlikely that real-world classes would consist of graphs with exactly the same variation for all factors. 
Therefore, it is not a freak occurrence that on average the distance between two graphs in one class sparsifies at a different rate than between those in another class. In fact, it is the opposite --- it would be quite a coincidence for graphs from two different classes to sparsify at exactly the same rate.
As a result, class-level density disparity as discussed in Sec. \ref{ssec:hypothesis} appears to be an inevitable consequence of using these types of representation learning techniques.

\section{Empirical Analysis}
\label{sec:exp}
To reiterate, we hypothesize that the performance flip issue occurs when samples from different classes have large enough density disparity and overlapping support. Provided those two conditions hold, outlier models tend to rank the class with sparser graph embedding above the denser class no matter which class is down-sampled at what rate. 
In this section, we present further measurement and quantitative analysis based on measures in Sec. \ref{ssec:measures} to support our hypothesis. Notice that we mainly provide measurement study for propagation based methods, due to their importance and easy-to-study mechanism. Sec. \ref{ssec:setup} introduce the setup in detail regarding datasets and model configurations. Sec. \ref{ssec:measure-flip} and Sec. \ref{ssec:measure-no-flip} contain measurement study of propagation based methods for one dataset with performance flip (DD) and the other without (IMDB), results for other datasets will be available at \projurl. Sec. \ref{sec:additional_result} provides a comprehensive performance study for 10 datasets and 11 detection models.


\subsection{Experiment Setup}\label{ssec:setup}

\begin{table}[h!]
\small 
    \caption{Dataset class semantics.}\label{tab:data-semantics}
    \centering
    \begin{tabular}{lccc}
    \toprule
    \textbf{Dataset} & \textbf{Domain} & \textbf{Class Labels} & \textbf{Class Semantic} \\
    \midrule  
    \multirow{2}{*}{DD \cite{dobson2003distinguishing}}  & Protein structure & 0 & Enzymes \\
                         & & 1 & Non-enzymes  \\
    \midrule 
    \multirow{2}{*}{PROTEINS \cite{borgwardt2005protein}}  & Protein structure  & 0 & Enzymes  \\
                                 & & 1 & Non-enzymes  \\
    \midrule
    \multirow{2}{*}{NCI1 \cite{wale2008comparison}}      & Molecular (drug)  & 0 & Inactive for anti-HIV   \\
                                 & & 1 & Active for anti-HIV     \\
    \midrule
    \multirow{2}{*}{IMDB-BINARY \cite{yanardag2015deep}} & Actors'   & 0 & Movie genre: Action  \\
                                 & collaboration network & 1 & Movie genre: Romance  \\   
    \midrule
    \multirow{2}{*}{Mutagenicity \cite{kazius2005derivation}} & Molecular & 0 & Mutagens  \\
                                 & & 1 & Not mutagens  \\  

    \midrule
    \multirow{2}{*}{AIDS \cite{riesen2008iam}} & Molecular & 0 & Active against HIV  \\
                                 & & 1 & Inactive against HIV  \\  
    \midrule
    ENZYMES \cite{schomburg2004brenda} & Protein structure & 0$\sim$5 & Different type of enzymes\\  
    \midrule
    REDDIT-5K  \cite{yanardag2015deep}& Discussion thread & 0$\sim$4 & \makecell{5 type of subreddits: worldnews,\\ videos, AdviceAnimals,\\aww and mildlyinteresting}\\
    \bottomrule                          
    \end{tabular}

\end{table}

\begin{table}[h!]
\small 
    \caption{Dataset summary statistics.} \label{tab:datastats}
    \centering
    \tabcolsep=0.085cm
    \begin{tabular}{lcccccc}
    \toprule
    Dataset   & Class & \#Graphs & \#Node Labels & Avg. \#Nodes & Avg. \#Edges & Avg. Degree \\ 
    \midrule
    \multirow{2}{*}{DD}  & 0 & 691 & 89 & 355.2 & 1806.6 & 5.04\\
                         & 1 & 487 & 89 & 183.7 & 898.8 &  4.88\\
    \midrule
    \multirow{2}{*}{PROTEINS}  & 0 & 663 & 3 & 50   & 188.1 & 3.79\\
                               & 1 & 450 & 3 & 22.9 & 83.1 &  3.64\\
    \midrule

    \multirow{2}{*}{NCI1}  & 0 & 2053 & 37 & 25.65 & 55.3 & 2.15\\
                           & 1 & 2057 & 37 & 34.07 & 73.9 & 2.17\\
    \midrule

    \multirow{2}{*}{IMDB}  & 0 & 500 & - & 20.1 & 193.5 & 9.1\\
                           & 1 & 500 & - & 19.4 & 192.5 & 8.6\\
    \midrule
    \multirow{2}{*}{Mutagenicity}  & 0 & 2401 & 14 & 29.3 & 60.5 & 2.1\\
                                   & 1 & 1936 & 14 & 31.4 & 62.7 & 2.0\\

    \midrule
    \multirow{2}{*}{AIDS}  & 0 & 400  & 38 & 37.6 & 80.5 & 2.13\\
                           & 1 & 1600 & 38 & 10.2 & 20.3 & 1.98\\
    \midrule
    \multirow{2}{*}{ENZYMES}  & 0 & 100 & 3 & 36.2 & 132.7 & 3.84\\
                              & 1 & 100 & 3 & 29.9 & 113.7 & 3.79\\
                              & 2 & 100 & 3 & 28.9 & 111.2 & 3.86\\  
                              & 3 & 100 & 3 & 38.2 & 148.8 & 3.99\\
    \midrule
    \multirow{2}{*}{REDDIT-5K}   & 0 & 1000 & - & 799.4 & 2035.5 & 2.52\\
                                 & 1 & 1000 & - & 852.1 & 1940.4 & 2.23\\
                                 & 2 & 1000 & - & 374.1 & 856.5 & 2.24\\  
                                 & 3 & 1000 & - & 249.6 & 534.0 & 2.11\\
                                 & 4 & 1000 & - & 267.0 & 581.7 & 2.14\\

    \bottomrule
    \end{tabular}
\end{table}

\textbf{Datasets.} We use 10 real-world binary labeled graph classification datasets where 4 of them are derived from 2 multi-class classification datasets (ENZYMES and REDDIT-5K) by picking 2 classes out. These 10 datasets come from three domains: molecular chemistry, social networks, and bioinformatics.  IMDB and REDDIT-5K are from social network domain, that contain unlabeled graphs, for which we create label by using their node degrees. We report their dataset statistics in Table \ref{tab:datastats}. We also provide detailed class semantics for every class in each dataset in Table \ref{tab:data-semantics}. Based on their class semantics we divide 10 datasets into 2 groups as mentioned in Sec.\ref{ssec:setup}: ``X\&Y'' type (datasets in Table \ref{tab:all_xy}) and ``X\&Non-X'' type (datasets in Table \ref{tab:all_xnonx}). ``X\&Non-X'' datasets contain one compact class ``X'' and the other broader class containing samples not in ``X'', while ``X\&Y'' datasets contain two compact/regular classes. We remark that even in ``X\&Non-X'' type dataset one may hardly decide on the natural, semantic outlier class -- e.g. in AIDS one can argue that being active against HIV should be outlier as it is rare and semantically important, yet one can also argue that in the context of developing new drugs for treating AIDS, detecting ineffective drug should be the outlier detection task. This suggests that in some cases repurposing graph classification datasets by down-sampling either class can be considered meaningful.



\noindent
\textbf{Model configuration of propagation based methods:} 
We provide both performance study and measurement study for propagation based methods: two-stage models WL+LOF/OCSVM and PK+LOF/OCSVM \footnote{PK and WL implementation use \cite{siglidis2020grakel}.}, as well as end-to-end deep-one-class model OCGIN. 
We study the behavior of these models under different number of propagations; 
specifically WL and PK iterations range from 1 to 11, and OCGIN embeddings are extracted from layers 0 (i.e. input node vectors) through 5.
For WL we specifically use WL subtree kernel. PK has a bin-width hyperparameter for hash function $\phi_{PK}$ (See Eq. \ref{eq:pk}), which is set to 0.1. Note that smaller bin-width leads to faster sparsification with a more severe performance flip. 
We use two different types of downstream outlier detectors: LOF (density-based) and OCSVM (one-class based). The LOF outlier detector is setup with default parameters ($k=$20 number of neighbors, and leaf size 30) from scikit-learn \cite{scikit-learn}. For OCSVM we use kernel-based SVM with the kernel output from graph kernels, and setup contamination factor $nu=0.1$. 
 
For OCGIN we use the default GIN implementation from \cite{conf/iclr/XuHLJ19}, where we remove bias terms at all layers to prevent feature collapse \cite{ruff2018deep}. A graph's representation is produced from the summation of all previous layers' hidden representations, with a mean pooling over all nodes in the graph. 
Note that we train OCGIN only on down-sampled variants of a dataset, as it is trained end-to-end assuming outliers to be minority. For figures utilizing full data, we simply feed-forward all the graphs in the database over the \textit{trained} model.
We set number of layers to $L=5$ and number of hidden units to 128 for all datasets. We use the Adam optimizer \cite{kingma2014adam} to train OCGIN with a $5\cdot 10^{-4}\  L_2$ penalty on weights. The model is trained for 25 epochs. All other hyperparameters are picked from typical/default values, since our goal is to illustrate the performance flip and related issues instead of achieving best performance. Hyperparameter selection for unsupervised deep outlier detection is an important problem which is outside the scope of this paper. 

\noindent

\textbf{Model configuration of non-propagation based methods:} 
We only provide measurement study for propagation based methods and omit the same for others as the underlying mechanism is different and harder to analyze. 
We use two graph embedding methods (Graph2Vec and FGSD \footnote{We use implementation in \cite{rozemberczki2020karate}.}) and 3 downstream outlier detectors (LOF, OCSVM, and Isolation Forest \cite{liu2008isolation}). Graph2Vec views graph as ``document'' and subgraph as ``word'' and learn the embedding using the Word2Vec \cite{mikolov2013efficient} algorithm. We use the default parameter implemented in KarateClub \cite{rozemberczki2020karate} with number of WL iterations set to 3. FGSD uses histogram of spectral distances among all node pairs to create graph embeddings and it only uses graph embedding without considering node labels. Similarly we use the default implementation given in KarateClub. As for three downstream outlier detectors, LOF and OCSVM share the same configuration with propagation based methods. Isolation Forest is not used for PK and WL as it does not support kernel matrix as input. We use the default setup in sklearn for it.

\noindent
\textbf{Pairwise similarity matrix.} Our proposed measures in Sec. \ref{ssec:measures} are computed on top of pairwise similarities among all graphs. For PK and WL kernels, the normalized kernel matrix is investigated. For OCGIN, where we only have access to graph embeddings, we calculate pairwise similarity as {(1 - normalized pairwise distance)} between two graphs,
using Euclidean distance (i.e. $L_2$ norm). Similarity matrix is 
normalized to range $[0,1]$ via dividing it by the largest element in the matrix. 

\noindent
{\bf Remark.} We (will) include all figures corresponding to those presented in the following as well as previous sections for all propagation based models on all datasets in project website (\projurl). Only a subset of them are presented in this manuscript for brevity.

\subsection{Measurement study: when performance flip occurs}\label{ssec:measure-flip}
We have observed performance flip on majority of datasets. In this section we present our measurement study and analysis based on DD dataset.  

\subsubsection{\bf Analysis on full data.} 
\begin{figure}[!h]
    \centering
    \includegraphics[width=\textwidth]{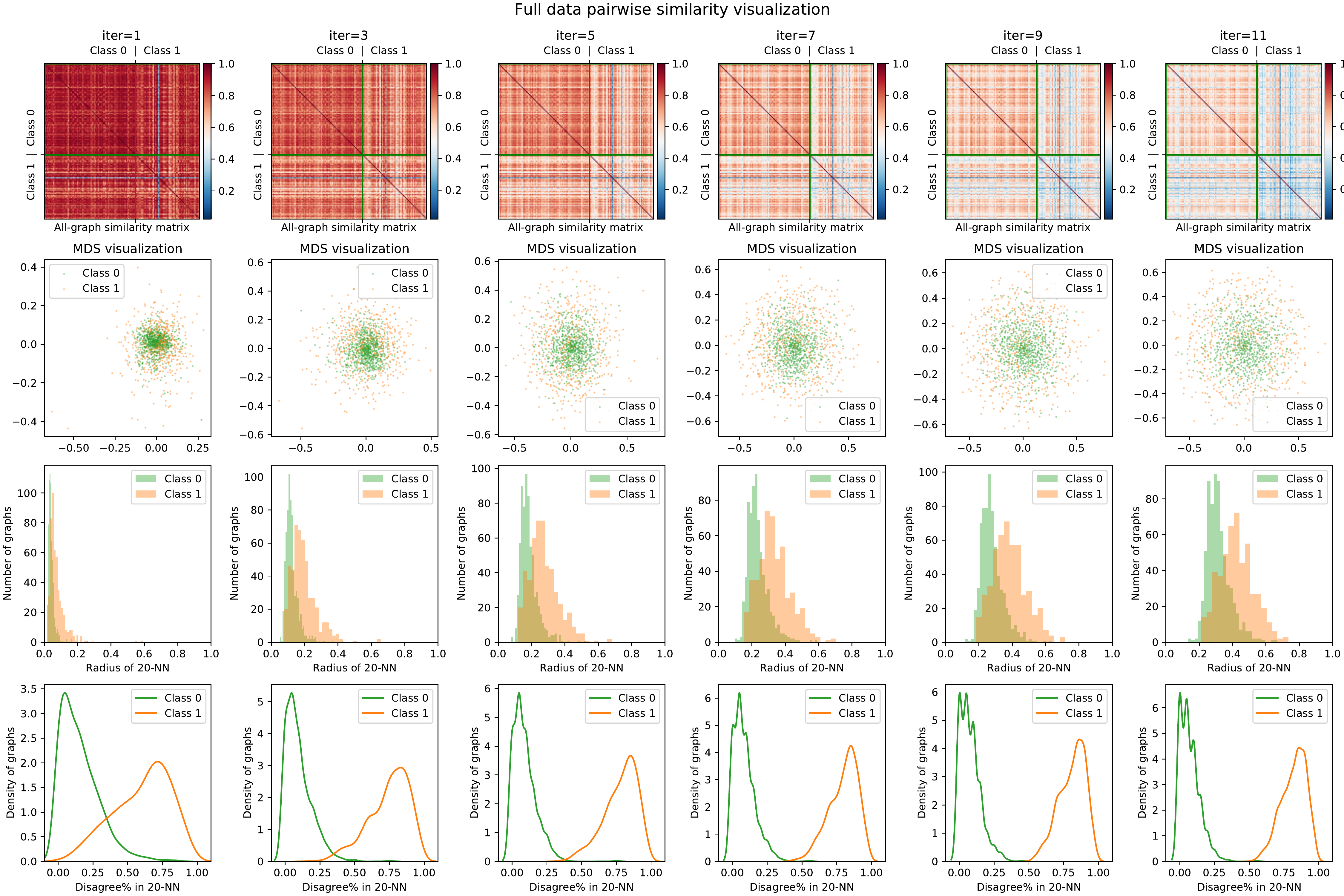}
    \caption{(Top row) Pairwise similarity matrix for all graphs in full DD dataset, based on WL subtree kernel over increasing iterations (left to right). (Second row) 2-d MDS visualization based on the similarity matrix. (Third row) Distribution of {\em NN-Radius} for all graphs in each class. (Last row) Distribution of {\em NN-Disagreement\%} for all graphs in each class.}\label{fig:DD-WL-fulldata}
\end{figure}

We start with analyzing the inherent differences between two classes regarding density on DD.
Fig. \ref{fig:DD-WL-fulldata} (top row) visualizes the all-pairs similarity matrix based on WL over increasing iterations where sparsification can be observed for both classes (left to right).
In the second row, 2-d MDS embeddings of all the graphs based on the corresponding pairwise similarities are shown.
Again sparsification can be visually confirmed based on the increasing spread of points (i.e. graphs) in each class.
In addition, we notice that class 0 (green points) \textit{sparsify faster} than class 1 (orange points), as the denser class 1 instances are surrounded by the dispersed class 0 instances.
We quantify this difference via the \rad measure, as in the third row, where the corresponding distributions of \rad for all graphs are shown for each class. Over iterations both class distributions shift to the right (i.e. sparsify). The shift is more evident for class 0 (i.e. speed of sparsification is larger) as the (green) histogram spreads out while the other (orange) histogram remains relatively peaked.

Next we analyze the overlapping support between two classes. The mixing of colors in the MDS visualizations is suggestive of overlap, however one can argue that it is an artifact of limited representation capacity of MDS in 2-d.
To quantify overlap more concretely, we show the distributions of \dis for all graphs in each class in the last row of Fig. \ref{fig:DD-WL-fulldata}.
 The distributions for class 1 concentrate more on the left side (majority of neighbors are from the \textit{same} class) whereas for class 0 they concentrate on the right (majority of neighbors are from the \textit{opposite} class).
 Both density disparity and overlapping support are increasingly more evident with increasing  number of iterations, which suggests that graph propagation amplifies the issue.



To summarize, this analysis confirms that class 0 in DD is the increasingly sparser class with larger {\em NN-Radius} (means sample density is smaller) and higher {\em NN-Disagreement\%} (means higher overlapping support). As a result, down-sampling class 0 as outlier induces a relatively `easy' task for WL+LOF. Moreover, the task becomes `easier' with more propagation as the graphs in class 0 spread out even further, hence the increasing performance gap.

\begin{figure}[!b]
	\centering
	\includegraphics[width=\textwidth]{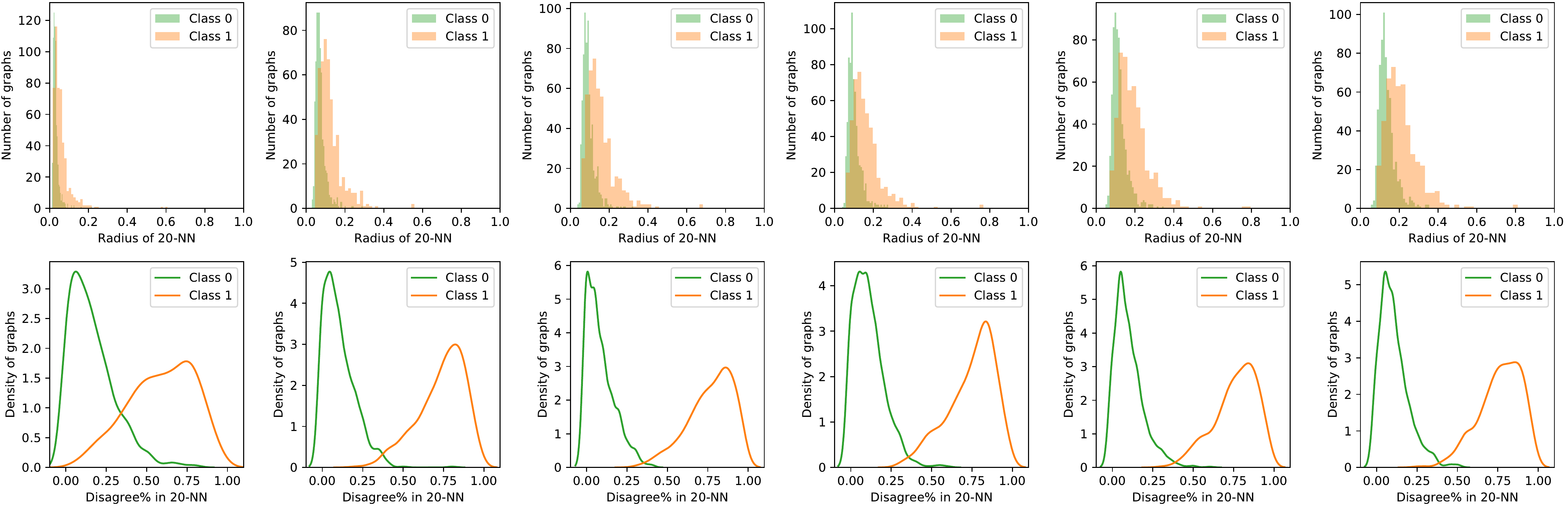}
	{\small{(a) PK}} 
	\includegraphics[width=\textwidth]{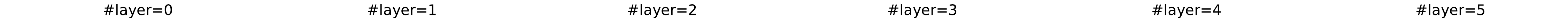}
	\includegraphics[width=\textwidth]{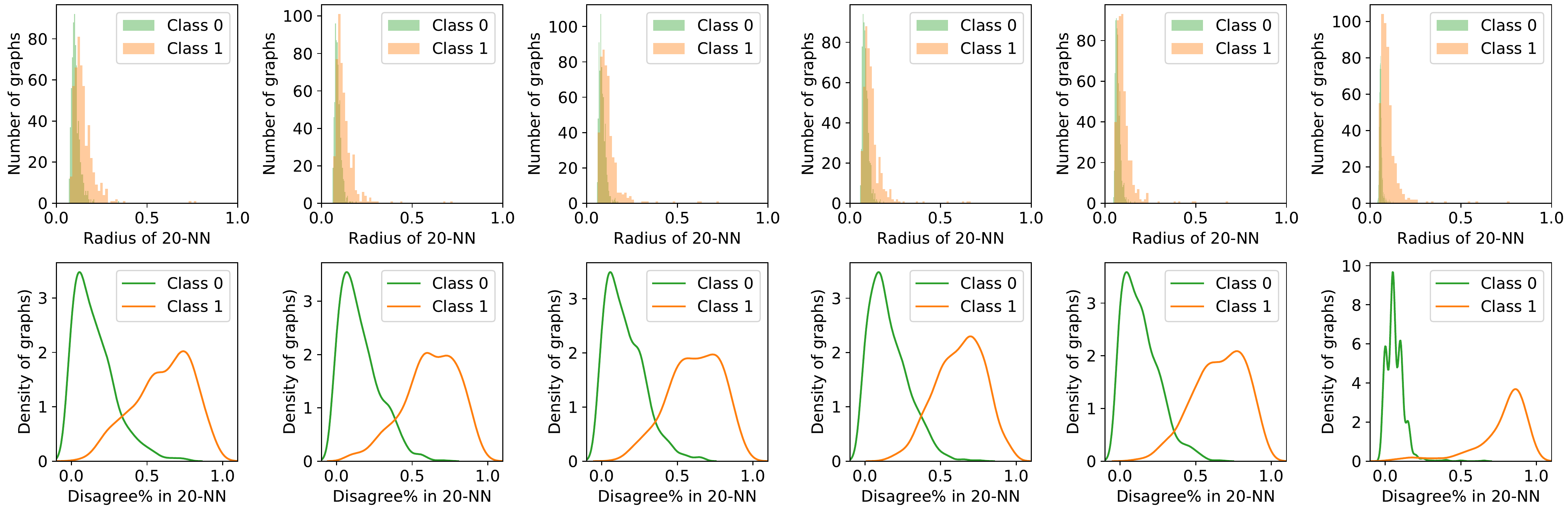}
	{\small{(b) OCGIN.}}
	\caption{Quantitative measures of density disparity and overlapping support on DD for (a) PK and (b) OCGIN for increasing graph propagation (left to right).}\label{fig:DD-PK-OCGIN-measures}
\end{figure}

The conclusions are similar for PK on DD, as shown in Fig. \ref{fig:DD-PK-OCGIN-measures}(a).
For OCGIN, while we continue to observe performance flip on DD variants, it is to a lesser degree. 
Specifically we find that the disparity between classes is not worsened with increasing graph propagation in this case, 
as suggested by Fig. \ref{fig:DD-PK-OCGIN-measures}(b).  In fact, the difference in distributions seems to close especially at the last layer.
Irrespectively, OCGIN performs considerably better when class 0 is down-sampled like the other models, meaning that it is not shielded from the performance flip issue that we identify.
We believe this is due to the initial disparity (See Fig. \ref{fig:DD-PK-OCGIN-measures}(b) at layer 0, i.e. original input), which it cannot recover from, despite model training. This is akin to a bad initialization (for one variant) of OCGIN  that ultimately leads to poor performance.

\subsubsection{\bf Analysis upon down-sampling.} 
\begin{figure}[!t]
	\centering
    \includegraphics[width=\textwidth]{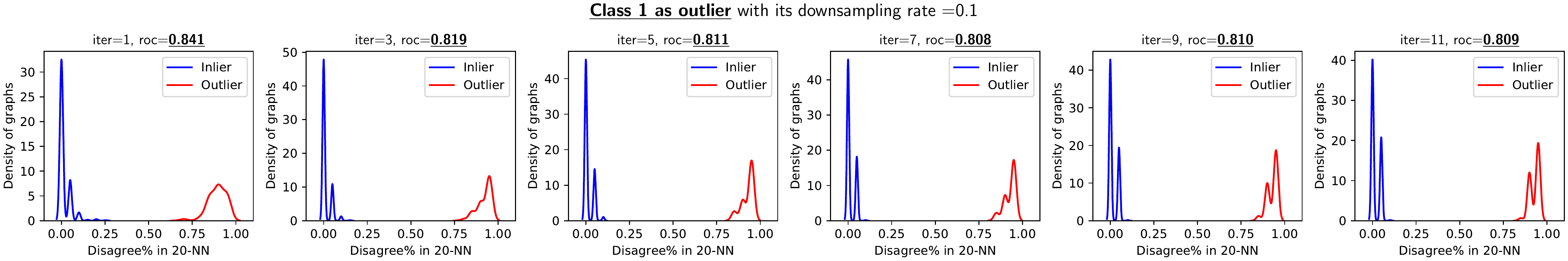}
    \includegraphics[width=\textwidth]{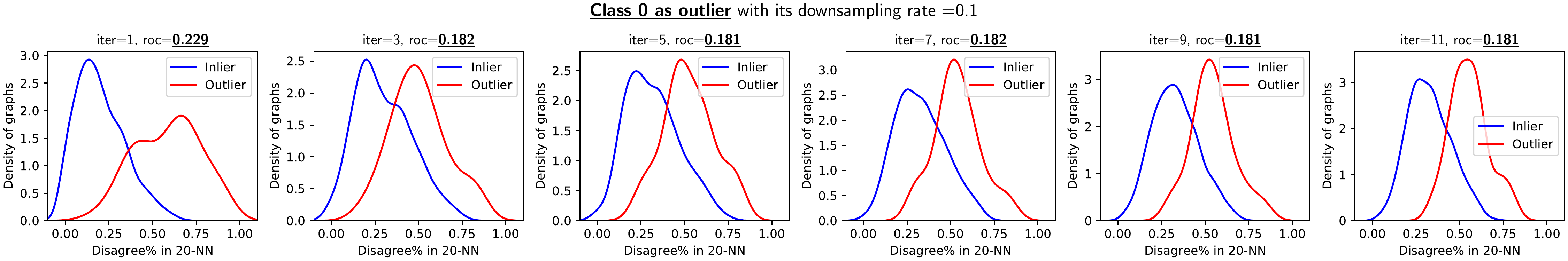}
	\caption{\dis distribution of outliers (top row: class 1 is down-sampled and bottom row: class 0 is down-sampled, in red) and inliers (vice versa, in blue) over WL iterations (left to right) on DD.}\label{fig:DD-WL-downsampled}
\end{figure}

Next we analyze the flip in performance upon down-sampling one class or the other via the contrast in {\em NN-Disagreement\%} distributions.  Fig. \ref{fig:DD-WL-downsampled}(top) and (bottom) respectively show those distributions when class 1 is down-sampled (denoted Outlier in red) and when class 0 is down-sampled (now denoted Outlier in red) for WL on DD.  
We notice the stark difference: At the (bottom), the inliers form a dense distribution with negligible mixing with the outliers. The outliers are dispersed, far from one another, as their NN-neighborhood mainly contains inliers. This is the `easy' detection task that well aligns with the underlying assumption of most outlier detectors -- hence the high peformance of WL+LOF.

In contrast, the (top) figures suggest that the inliers and outliers are intermixed, which worsens with propagation as the distributions become more and more indistinguishable. (Note that down-sampled class inherently has a right-shifted distribution as down-sampling induces sparsity.)
This corresponds to the `hard' detection task where it is difficult to distinguish inliers from outliers -- hence the poor, in fact worse-than-random performance.


\begin{figure}[!h]
	\centering
	\begin{tabular}{c}
    \includegraphics[width=\textwidth]{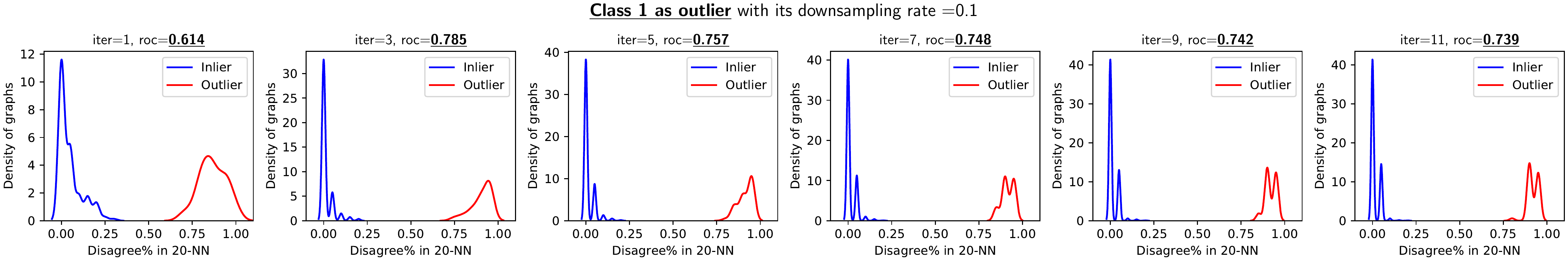}\\
    \includegraphics[width=\textwidth]{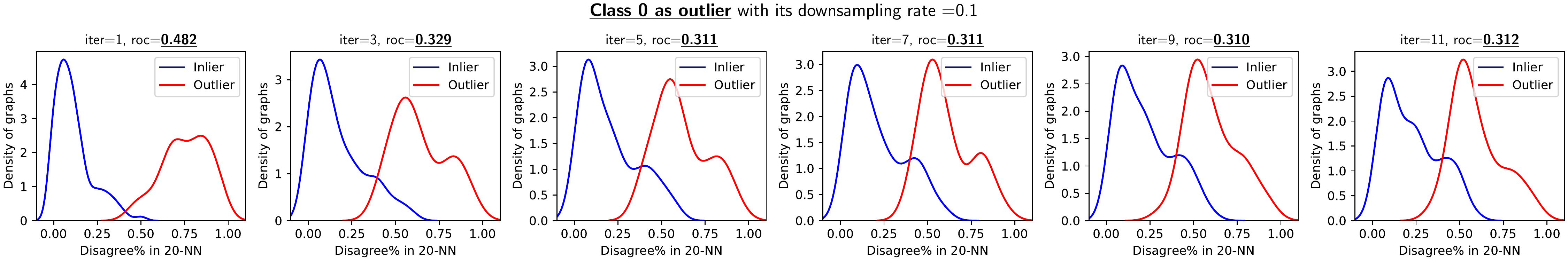}\\

	{\small{(a) PROTEINS}} \\\\ 
    \includegraphics[width=\textwidth]{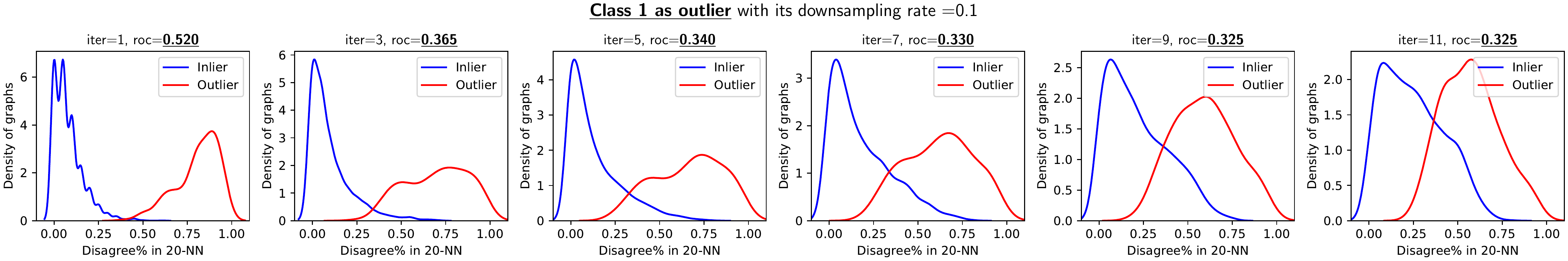}\\
    \includegraphics[width=\textwidth]{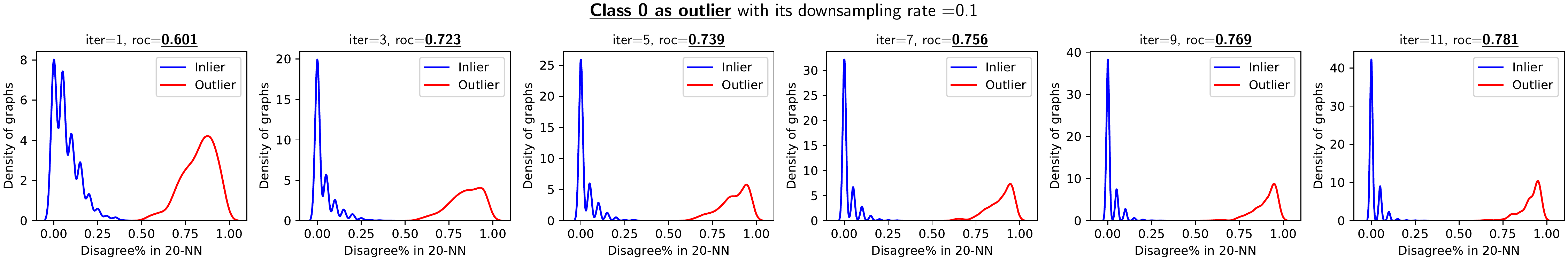}\\
	{\small{(b) NCI1}}
	\end{tabular}
	\caption{\dis distribution of outliers (top rows: when class 1 is down-sampled and bottom rows: when class 0 is down-sampled, in red) and inliers (vice versa, in blue) over WL iterations (left to right) on (a) PROTEINS and (b) NCI1.} \label{fig:DD-PROTEINS-NCI1-measures}
\end{figure}

The conclusions are similar for WL+LOF on PROTEINS and NCI1, as shown in Fig. \ref{fig:DD-PROTEINS-NCI1-measures}(a) and (b), respectively (except for NCI1 down-sampling class 0 happens to induce the `hard' task).
In fact, performance appears to be inversely correlated with the amount of overlap between the \dis distributions of inliers and outliers. We refer to the Appendix in \cite{zhao2020using} for similar results on these three datasets for PK+LOF and OCGIN.

\begin{figure}[!t]
	\centering
	\includegraphics[width=\textwidth]{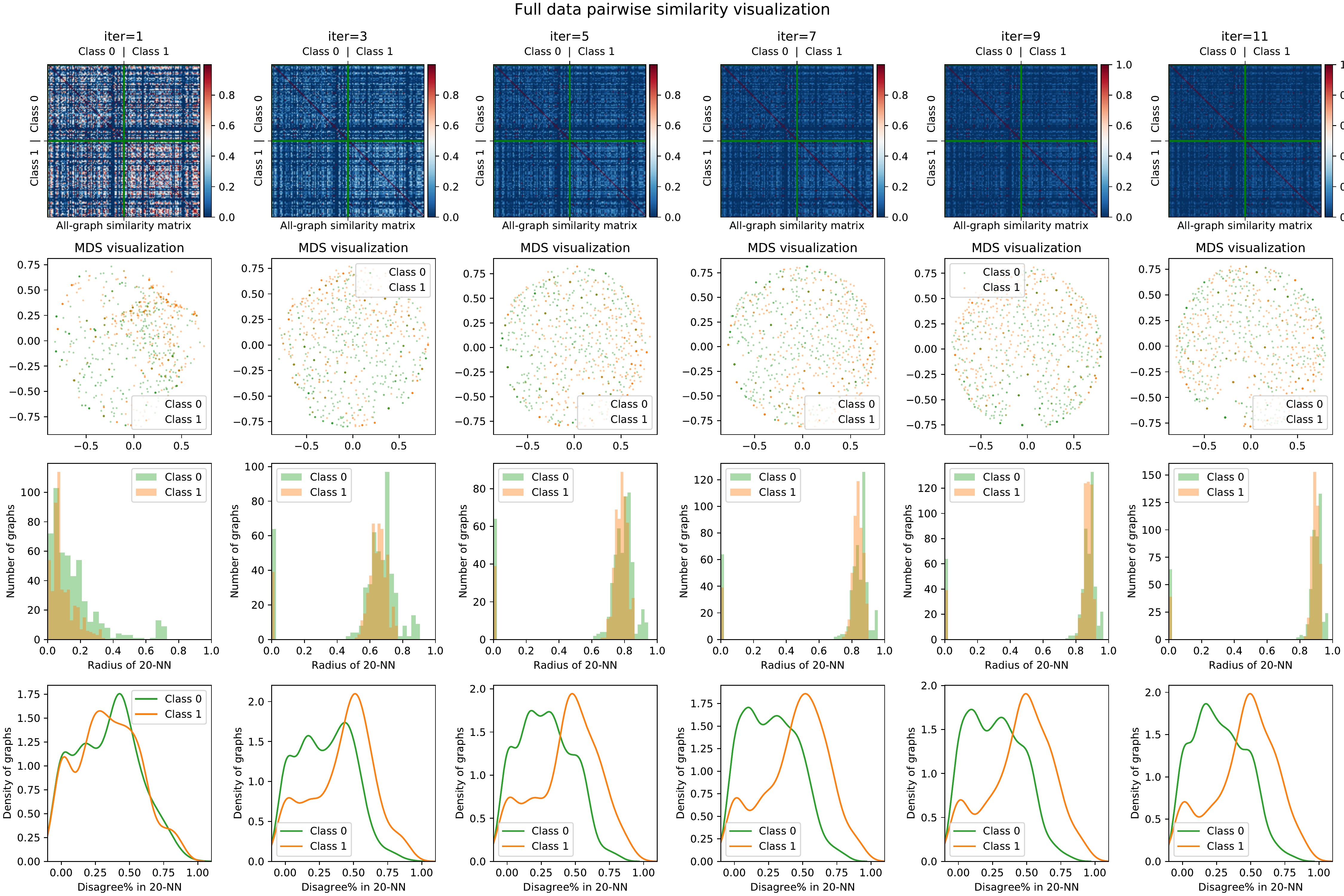}
	\caption{(Top row) Pairwise similarity matrix for all graphs in full IMDB-BINARY dataset, based on WL subtree kernel over increasing iterations (left to right). (Second row) 2-d MDS visualization based on the similarity matrix. (Third row) Distribution of {\em NN-Radius} for all graphs in each class. (Last row) Distribution of {\em NN-Disagreement\%} for all graphs in each class.}\label{fig:IMDB-BINARY-WL-fulldata}
\end{figure}

\begin{figure}
	\centering
	\includegraphics[width=\textwidth]{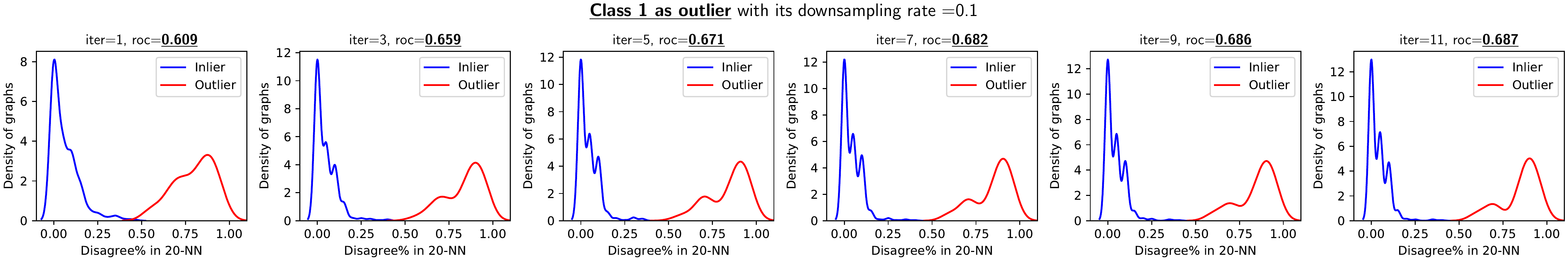}\\
	\includegraphics[width=\textwidth]{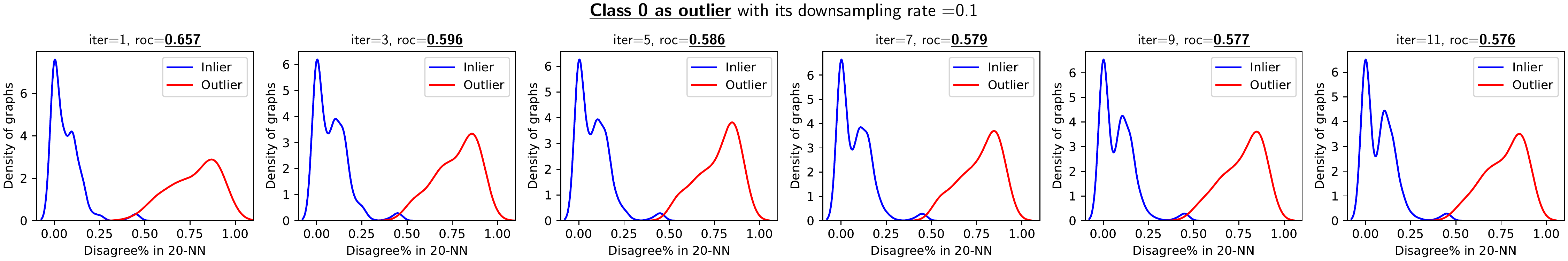}
	\caption{\dis distribution of outliers (top row: class 1 is down-sampled and bottom row: class 0 is down-sampled, in red) and inliers (vice versa, in blue) over WL iterations (left to right) on IMDB-BINARY.
	}\label{fig:IMDB-BINARY-WL-downsampled}
\end{figure}

\subsection{Measurement study: when performance flip does not occur}\label{ssec:measure-no-flip}

Although the performance flip issue occured on a considerable number of datasets we have experimented with,
it is not always observed. In this section, we present a similar analysis for IMDB-BINARY for comparison purposes. Results for other datasets and propagation based methods will be available at \projurl.

As shown in Fig. \ref{fig:IMDB-BINARY-WL-fulldata}, sparsification arises rather fast on IMDB-BINARY with increasing WL iterations -- notice the drastic shift of the \rad distributions from left-most to right-most (third row). This can also be visually confirmed from the pairwise similarity matrix (top row) as the heatmap gets darker blue (from left to right).
Interestingly, the rate (or speed) of sparsification appears to be similar among the two classes.
As such, density disparity does not seem to arise -- notice the similar mixing among the colored points in 2-d MDS visualization (second row). On the other hand, we continue to observe the overlapping support (i.e. mixing) of graph embeddings to a large extent (last row). 
 
 These suggest that down-sampling any one of the classes as outlier would induce two detection tasks with similar difficulty.
 Fig. \ref{fig:IMDB-BINARY-WL-downsampled} confirms this hypothesis, where the distribution of inliers and outliers in the embedding space look similar between the two down-sampled variants of IMDB-BINARY.
 Moreover, the outliers are sparser and more dispersed as compared to the inliers, which consequently leads to  better-than-random performance on both tasks. The ROC-AUC values are not too high due to the mixing of the embeddings that makes the detection task harder. In Sec. \ref{ssec:hypothesis} we argued that density disparity alone is not a sufficient condition for performance flip. This result on IMDB-BINARY shows that overlapping support (i.e. mixing) alone is also not a sufficient condition for the performance flip. Both conditions together give rise to the issue.

To conclude, our analysis sheds light onto the leading factors (density disparity and overlapping support) as well as contributing factors (graph propagation) behind the performance flip issue. These help us reason about both cases in which it occurs or not. However, we do not have a full understanding of why these factors arise for some datasets but not the others in the first place. Future research could focus on studying this disparity between datasets. 

\hide{
\begin{table}[h!]
    \caption{Average ROC-AUC performance for OCGIN}
    \centering
    \begin{normalsize}
    \begin{tabular}{lccccccc}
    \toprule
     Dataset & Outlier Class & \#L=0 & \#L=1 & \#L=2 & \#L=3 & \#L=4 & \#L=5  \\
    \midrule
    \multirow{2}{*}{DD epoch=0} & 0& 0.718 &   0.552 &   0.556 &   0.543 &   0.573 &   0.611\\
                                & 1& 0.339 &   0.426 &   0.384 &   0.382 &   0.377 &   0.376\\
    \multirow{2}{*}{DD epoch=25} & 0& 0.712 &   0.797 &   0.751 &   0.754 &   0.760 &   0.771\\
                                 & 1& 0.318 &   0.265 &   0.298 &   0.313 &   0.334 &   0.352\\
    \multirow{2}{*}{DD epoch=25 [s]} & 0& 0.781 &   0.756 &   0.732 &   0.729 &   0.731 &   0.771\\
                                     & 1& 0.278 &   0.290 &   0.317 &   0.338 &   0.344 &   0.352\\
    \midrule
    \multirow{2}{*}{PROTEINS epoch=0}  & 0& 0.729 &   0.776 &   0.767 &   0.746 &   0.722 &   0.690\\
                                       & 1& 0.305 &   0.275 &   0.266 &   0.287 &   0.310 &   0.354\\
    \multirow{2}{*}{PROTEINS epoch=25} & 0& 0.740 &   0.754 &   0.731 &   0.724 &   0.704 &   0.685\\
                                       & 1& 0.319 &   0.298 &   0.366 &   0.355 &   0.399 &   0.375\\ 
    \multirow{2}{*}{PROTEINS epoch=25[s]} & 0& 0.714 &   0.754 &   0.772 &   0.782 &   0.784 &   0.685\\
                                          & 1& 0.342 &   0.312 &   0.295 &   0.292 &   0.291 &   0.375\\ 
    \midrule
    \multirow{2}{*}{NCI1 epoch=0} & 0& 0.462 &   0.489 &   0.506 &   0.501 &   0.491 &   0.472\\ 
                                  & 1& 0.545 &   0.529 &   0.513 &   0.520 &   0.520 &   0.521\\
    \multirow{2}{*}{NCI1 epoch=25} & 0& 0.440 &   0.459 &   0.463 &   0.467 &   0.471 &   0.453\\
                                   & 1& 0.583 &   0.592 &   0.596 &   0.647 &   0.646 &   0.653\\                            
    \multirow{2}{*}{NCI1 epoch=25[s]} & 0& 0.465 &   0.471 &   0.486 &   0.493 &   0.502 &   0.453\\
                                      & 1& 0.548 &   0.549 &   0.545 &   0.543 &   0.535 &   0.653\\     
    \midrule
    \multirow{2}{*}{IMDB epoch=0} & 0& 0.535 &   0.603 &   0.608 &   0.640 &   0.668 &   0.686\\
                                  & 1& 0.531 &   0.357 &   0.465 &   0.413 &   0.337 &   0.327\\
    \multirow{2}{*}{IMDB epoch=25} & 0& 0.587 &   0.659 &   0.512 &   0.496 &   0.500 &   0.492\\
                                   & 1& 0.551 &   0.567 &   0.654 &   0.678 &   0.649 &   0.673\\
    \multirow{2}{*}{IMDB epoch=25[s]} & 0& 0.521 &   0.477 &   0.452 &   0.465 &   0.472 &   0.492\\
                                      & 1& 0.525 &   0.536 &   0.555 &   0.576 &   0.628 &   0.673\\                                   
    \bottomrule
    \end{tabular}
    \end{normalsize}
\end{table}
}

\subsection{Performance study: All GLOD methods} \label{sec:additional_result}

In the previous subsections, we have demonstrated an understanding and analysis over propagation based GLOD methods. It is natural to ask whether only propagation based methods suffer from performance flip. Some observations are already summarized in Sec.\ref{sec:method}. FGSD operates in the Laplacian eigen-space, and Graph2Vec employs skip-gram based training via negative sampling, etc. As such, these other methods are harder to understand and analyze in detail. 
In this section we present a full performance study over 10 datasets and these additional methods to empirically answer the following questions:
\bit
	\item Q1: Is performance flip only restricted to propagation based methods? \label{q1}
	\item Q2: Do semantics of the dataset have any influence on performance flip? \label{q2}
	\item Q3: For two-stage methods (with first embedding and then outlier detection stages), how does each stage affect performance flip? \label{q3}
	\item Q4: Does the end-to-end method have advantage over two-stage methods? \label{q4}
\eit
We answer these questions sequentially in the following sections A1 (Sec. \ref{ssec:a1}), A2 (Sec. \ref{ssec:a2}), A3 (Sec. \ref{ssec:a3}), A4 (Sec. \ref{ssec:a4}). 
Besides observations, we provide our understanding and hypothesis. 
Finally, we point out three key questions that we believe are important to GLOD task (Sec. \ref{ssec:questions}). 
As discussed before, we divide the datasets into ``X\&Non-X'' and ``X\&Y'' type. We evaluate propagation based two-stage methods (WL, PK), end-to-end propagation based method (OCGIN), and graph embedding based two-stage methods (Graph2Vec, FGSD) for all datasets. Table \ref{tab:all_xnonx} shows all the results on ``X\&Non-X'' type datasets, and Table \ref{tab:all_xy} contains all results on ``X\&Y'' type datasets. We carefully analyze these results to answer the above questions next.

\bgroup
\begin{table}
\centering
\footnotesize        
\captionsetup{font=small}
    \caption{Average ROC-AUC (over 10 random seeds) of 5 different graph embedding methods and 3 different outlier detectors over 5 ``X\&Non-X '' type datasets. Each dataset has 2 down-sampled variants. `DC' stands for down-sampled class, which is also outlier class. Cells colored  with \textcolor{red!60}{Red}, \textcolor{green!60}{Green}, \textcolor{yellow!100}{Yellow} represent: performance of both variants are worse than random, both variants are better than random, and performance flip scenario, respectively. Performance flip is widely observed. Among all cases, \textbf{67.3\%} have performance gap$\ge0.2$, \textbf{52.7\%} cases have performance gap $\ge0.3$, \textbf{30.9\%} have performance gap$\ge0.4$. (G2V=Graph2Vec)}\label{tab:all_xnonx}
    \centering
    \begin{tabular}{l|cc|cc|cc|cc|cc}
    \toprule
    Methods & \multicolumn{2}{c|}{DD} & \multicolumn{2}{c|}{PROTEINS} & \multicolumn{2}{c|}{NCI1} & \multicolumn{2}{c|}{Mutagenicity} & \multicolumn{2}{c}{AIDS} \\
    \cline{2-11}
        & DC=0 & DC=1 & DC=0 & DC=1 & DC=0 & DC=1 & DC=0 & DC=1  & DC=0 & DC=1\\
        & ``X'' & ``Non-X'' & ``X'' & ``Non-X'' & ``Non-X''& ``X''& ``X''& ``Non-X''  & ``X''& ``Non-X''\\
     \midrule
	 WL-LOF &  \flip 0.186 &  \flip 0.815 & \flip  0.276 & \flip  0.664 &  \flip 0.730 & \flip  0.349  &  \flip 0.460 &  \flip 0.629 & \flip 0.193 & \flip 0.950\\
	 PK-LOF &  \flip 0.194 &  \flip 0.824 & \flip  0.389 & \flip  0.557 &  \flip 0.678 & \flip  0.366  &  \flip 0.480 &  \flip 0.613 & \flip 0.387 & \flip 0.896\\
	 WL-OCSVM &  \flip 0.179 &  \flip 0.820 & \flip  0.189 & \flip  0.794 & \flip  0.653 & \flip 0.341 & \good  0.500 & \good 0.540 & \flip 0.048 & \flip 0.972\\
	 PK-OCSVM &  \flip 0.222 &  \flip 0.809 & \flip  0.244 & \flip 0.751 &  \flip 0.593 &  \flip 0.429 & \good  0.517 & \good 0.541 & \flip 0.175 & \flip 0.880\\
	 OCGIN-5  &  \flip 0.327 &  \flip 0.720 & \flip  0.370 & \flip 0.681 &  \flip 0.643 &  \flip 0.467 & \good  0.503 & \good 0.650 & \flip 0.200 & \flip 0.922\\
	 \midrule
	 \midrule
	 G2V-LOF & \flip  0.680 & \flip  0.362 & \flip  0.644 & \flip  0.414 &  \good 0.594 & \good  0.588  &  \flip 0.495 & \flip 0.616 & \flip 0.919 &\flip 0.424\\
	 FGSD-LOF      & \flip  0.628 & \flip  0.425 & \bad   0.468 & \bad   0.422 &  \flip 0.712 & \flip  0.417  &  \flip 0.458 & \flip 0.634 & \flip 0.934 &\flip 0.290\\
	 G2V-OCSVM & \flip  0.631 & \flip  0.336 & \flip  0.569 & \flip  0.466 & \flip  0.332 & \flip  0.634  & \bad  0.492 & \bad 0.491 & \flip 0.903 & \flip 0.035\\
	 FGSD-OCSVM      & \flip  0.384 & \flip  0.781 & \flip  0.385 & \flip  0.711 & \good  0.545 &  \good 0.550  & \flip 0.366 & \flip 0.664 & \good 0.979 & \good 0.743\\
	 G2V-IF    & \flip  0.656 & \flip  0.335 & \flip  0.552 & \flip  0.449 & \flip  0.341 & \flip  0.636  & \flip 0.446 & \flip 0.586 & \flip 0.904 & \flip 0.035\\
	 FGSD-IF         & \flip  0.745 & \flip  0.400 & \flip  0.773 & \flip  0.272 & \flip  0.384 & \flip  0.637  & \flip 0.476 & \flip 0.569 & \flip 0.984 & \flip 0.018\\
	\bottomrule                        
    \end{tabular}

    \caption{Same configuration as Table \ref{tab:all_xnonx}, this time over 5 ``X\&Y '' type datasets. FGSD cannot run over REDDIT datasets. Performance flip is not observed for IMDB dataset across all methods. OCGIN has performance above random across all datasets. Performance flip is still widely observed for other methods. Among all cases, \textbf{30.6\%} have performance gap$\ge0.2$, \textbf{22.4\%} cases have performance gap $\ge0.3$, \textbf{12.2\%} have performance gap$\ge0.4$.(G2V=Graph2Vec)}\label{tab:all_xy}
    \centering
    \begin{tabular}{l|cc|cc|cc|cc|cc}
	\toprule
	Methods & \multicolumn{2}{c|}{IMDB-BINARY} & \multicolumn{2}{c|}{ENZYMES(c0\&c1)} & \multicolumn{2}{c|}{ENZYMES(c2\&c3)} & \multicolumn{2}{c|}{REDDIT(c0\&c1)} & \multicolumn{2}{c}{REDDIT(c2\&c3)}\\
	\cline{2-11}
	       & DC=0 & DC=1  & DC=0 & DC=1 & DC=2 & DC=3 & DC=0 & DC=1 & DC=2 & DC=3\\
	\midrule
	WL-LOF   & \good 0.603 & \good 0.651&  \good 0.518 &  \good 0.519 &  \flip 0.758 & \flip  0.399 &  \flip 0.261 & \flip  0.734 & \flip 0.189 & \flip  0.810\\
	PK-LOF   & \good 0.624 & \good 0.581&  \good 0.553 &  \good 0.550 &  \good 0.755 &  \good 0.633 &  \flip 0.485 & \flip  0.579 & \flip 0.361 & \flip  0.640\\
	WL-OCSVM & \good 0.524 & \good 0.571&  \flip 0.598 & \flip  0.385 & \flip  0.607 & \flip  0.462 & \flip  0.268 & \flip  0.739 & \flip 0.180 & \flip  0.821\\
	PK-OCSVM & \good 0.538 & \good 0.548&  \flip 0.590 & \flip  0.402 & \flip  0.456 & \flip  0.642 & \flip  0.388 & \flip  0.696 & \flip 0.207 & \flip  0.802\\
	OCGIN-5  & \good 0.643 & \good 0.508&  \good 0.615 & \good  0.517 & \good  0.587 &  \good 0.636 & \good  0.662 & \good  0.622 & \good 0.573 &  \good 0.608\\
	\midrule
	\midrule
	G2V-LOF   & \good 0.534 & \good 0.558& \flip  0.551 & \flip  0.354 & \flip  0.398 & \flip  0.532 & \flip 0.376 & \flip 0.563 & \flip 0.678 & \flip 0.323\\
	FGSD-LOF        & \good 0.606 & \good 0.505& \flip  0.644 &  \flip 0.412 & \flip  0.449 & \flip  0.608 &  -  &   - &     -  &  -\\
G2V-OCSVM & \good 0.526 & \good 0.551& \flip  0.565 & \flip  0.419 & \flip  0.492 & \flip  0.649 & \flip 0.465 & \flip 0.528 & \flip 0.679 & \flip 0.329\\
	FGSD-OCSVM      & \good 0.516 & \good 0.586& \good  0.531 & \good  0.503 & \good  0.523 & \good  0.613 &   - &   - &   - & -\\
	G2V-IF    & \good 0.516 & \good 0.562& \flip  0.572 & \flip  0.346 & \flip  0.410 & \flip  0.604 & \good  0.505 & \good  0.545 & \flip  0.695 & \flip  0.326\\
	FGSD-IF         & \good 0.522 & \good 0.617& \flip  0.537 & \flip  0.456 & \flip  0.300 & \flip  0.708 &   - &   - &   - & -\\
	\bottomrule                        
    \end{tabular}
\end{table}
\egroup

\hide{

Observations and understanding:
\begin{itemize}
  \item For ``X\&Non-X '' type datasets, performance flip is widely observed across all graph embedding methods and downstreaming outlier detectors. Propagation based methods agree on which version of the downsample achieving higher ROC-AUC. However in general, which downsample version having high performance depends on the choice of graph embedding method and outlier detector. As we only understand the mechanism of propagation based methods, we can only analyze it in detail. 
  \item For ``X\&Y '' type datasets, performance flip is widely observed for all methods except OCGIN. As OCGIN is learning based method that has ability to capture information of majority class, it achieves better-than-random performance for both downsample versions of all ``X\&Y '' type datasets. However, this ability of capturing distribution of majority class is not powerful enough to overcome the ``X\&Non-X '' type dataset bias. 
  \item Interestingly, although performance flip is widely observed, the graph embedding methods we used (FGSD and Graph2Vec) have mostly reversed pattern on which downsample version having better performance, comparing with propagation based methods.
  \item Above observation suggests the performance flip is not a dataset based problem as the downsample version with semantic anomaly doesn't consistently perform high across methods. Instead, it's more likely due to the misalignment between embedding generated by unsupervised graph embedding method and the assumption of outlier detector. 
  \item Unsupervised graph embedding methods generate embedding deterministically resulting that embeddings of one class be consistently denser and the other sparser in the embedding space, and downsampling the class with denser embeddings leads to worse-than-random ROC-AUC for most outlier detectors. Learning based GNN method has certain level ability of capturing majority class distribution but the propagation based inductive bias makes it harder for ``X\&Non-X '' type datasets.
\end{itemize}
}

\subsubsection{A1: Performance flip occurs across all methods.}\label{ssec:a1}
To help summarize results we use \textcolor{yellow!100}{\bf yellow} color to mark performance flip in Table \ref{tab:all_xnonx} and Table \ref{tab:all_xy}. The dominance of yellow for both propagation based (first row-wise block) and non-propagation based methods (second row-wise block) empirically verify that performance flip is not restricted to propagation based methods but arises more generally for other methods as well. 

The observation strengths the prevalence of performance flip for GLOD task, and raises the question of how to evaluate models for GLOD if the issue is persistent across datasets and models.

\subsubsection{A2: Dataset semantics play a role in performance flip.}\label{ssec:a2}
Clearly, the percentage of yellow cells in Table \ref{tab:all_xnonx} is noticeably larger than that in  Table \ref{tab:all_xnonx}, which suggests that \textbf{(1)} performance flip occurs \textbf{more often} in ``X\&Non-X '' type datasets than ``X\&Y '' type datasets. We further investigate the performance gap (i.e. severity of performance flip) in both tables numerically. Table \ref{tab:all_xnonx} has 67.3\% cases with performance gap $\ge$ 0.2, 52.7\% cases with performance gap $\ge$ 0.3, and 30.9\% cases with performance gap $\ge$ 0.4. In contrast, Table \ref{tab:all_xy} has only 30.6\% cases with performance gap $\ge$ 0.2, 22.4\% cases with performance gap $\ge$ 0.3, and 12.2\% cases with performance gap $\ge$ 0.4. The dramatic difference of performance gap between two types of datasets suggests that \textbf{(2)} performance flip applies \textbf{more severely} to ``X\&Non-X '' type datasets than ``X\&Y '' type datasets.  Finally, we also investigate which version of the downsample achieves high ROC-AUC. In Table \ref{tab:all_xnonx} we annotate class labels 0 and 1 with corresponding ``X'' and ``Non-X'' based on class semantics, where ``X'' refers to the compact class. Interestingly, for propagation based methods, down-sampling ``Non-X'' class as outlier class consistently achieves high performance while down-sampling ``X'' has performance worse than random. We conclude that \textbf{(3)}  downsampling the class with diverse patterns creates sparser, dispersed outliers in the embedding space when using propagation based methods, resulting in an easier outlier detection task with high ROC-AUC. 

We remark that the correlation between class semantic and performance flip challenges whether we should repurpose $X\&Non-X$ type graph classification dataset to evaluate GLOD. Further, one may argue we can always down-sample class ``Non-X'' as outlier to evaluate, however the ``Non-X'' class may not be a semantic outlier class (see Mutagenicity where ``Non-X'' represents non-mutagen that should be normal class in semantic), and for other non-propagation based methods the performance is not always high (see AIDS).  

\subsubsection{A3: Both embedding method and outlier detector affect performance flip.}\label{ssec:a3}
Previously we have observed that the propagation based methods consistently achieve high ROC-AUC when ``Non-X'' is down-sampled as outlier in ``X\&Non-X'' datasets. However this is not always true for other methods and datasets. We observe that both graph embedding and downstream outlier detector affect which version of downsample has high performance. See for example the AIDS dataset in Table \ref{tab:all_xnonx}, where  using different graph embedding methods has dramatically reversed performance flip pattern. Also see the DD dataset where FGSD+OCSVM and FGSD+LOF have reversed performance flip. The observation can be explained by the driving mechanism identified in Sec.\ref{ssec:hypothesis}. The performance of two-stage methods is determined by the (mis)alignment of embedding space generated by graph embedding methods as well as the assumptions of downstream outlier detector. In that sense, both graph embedding and outlier detection methods play a role in whether performance flip occurs and the amount of performance gap. 

\subsubsection{A4: End-to-end method can partially capture distribution of majority.}\label{ssec:a4}
All two-stage methods we studied are unsupervised and produce deterministic embeddings regardless of dataset distribution. This is not ideal for outlier detection, as they do not have enough ability to generate embeddings that capture distribution of the majority class. On the other hand, end-to-end methods learn from the input dataset by optimizing a loss (e.g. reconstruction error or distance to center) contributed by each sample in the dataset, and should have certain degree of ability to capture the majority (by achieving lower average loss for the majority class). As we can see in Table \ref{tab:all_xy}, OCGIN is the only method that does not  have performance flip for all 5 ``X\&Y'' datasets. This empirically shows certain ability of capturing majority class for OCGIN. 
Nevertheless, performance flip still occurs widely for OCGIN in all ``X\&Non-X'' datasets. Our hypothesis is that although OCGIN has certain ability to capture majority, the inductive bias (underlying mechanism) shared across all propagation based methods is too strong on ``X\&Non-X'' datasets to be corrected by learning.

The relatively higher robustness of our end-to-end method, OCGIN, against performance flip, especially on ``X\&Y'' type datasets, appears to be a promising direction for future work. This also motivates further analysis of other end-to-end GLOD methods.

\subsubsection{Three key questions for GLOD}\label{ssec:questions}
Based on our observations over all datasets and methods, we pinpoint three important questions for GLOD and aim to draw the community's attention toward answering them:
\bit
    \item Given the widely observed performance flip issue across models and datasets, what is the best way to evaluate detectors when repurposing graph classification datasets? 
    Is it fair to compare methods based on their average performance across all datasets, as one would usually do, while we now expect that some downsample versions will likely yield significantly worse-than-random performance?   
    \item As graph embedding method play a significant role on GLOD, how can we design better graph-level embedding methods (either unsupervised or end-to-end) to overcome performance flip issue specifically, and achieve better GLOD performance overall at large?    
     \item Given that methods do not agree on which version of downsample yields high performance, given any new detection task (i.e. dataset) and a set of candidate GLOD methods,
    how can we design unsupervised model selection strategies that can effectively choose a model that achieves high detection performance?
\eit
We believe these questions stand on the foundation of GLOD and provide solid starting points for the community toward progress on GLOD problems.

\section{Conclusion}
\label{sec:discuss}
\subsection{Summary}

Many researchers and practitioners repurpose binary classification datasets for outlier detection via down-sampling one of the classes to constitute the outliers. In graph-level outlier detection, we found that most binary classification datasets, when used in this fashion, introduce class-level
bias for different variants of downsample, resulting in dramatically different outlier detection performance. In this paper, we call this phenomenon ``performance flip''. 
This bias exhibits itself as disparity in detection task difficulty determined by the alignment of graph embedding density and assumption of downstream outlier detector. Looking over embedding space, the issue stems from two factors: (1) density disparity; one class being more compact than the other in the embedding space; and (2) overlapping support; mixing of the embeddings between classes. Given that most outlier detectors assume a compact set of inliers and scattered outliers, down-sampling the class that is sparse in the embedding space induces such a scenario with scattered outliers and hence achieves high detecting performance. In contrast, down-sampling the other class with denser embeddings induces clustered outliers (that shares support with normal class) and results in significantly worse-than-random performance.

Under unsupervised learning, the embedding space is determined by two parts: the input dataset and the embedding method. These two parts can interact with each other. Categorizing datasets into two as 'X\&Non-X' and 'X\&Y' reveals that 'X\&Non-X' type datasets have higher probability of performance flip with more severe performance gap. Empirical results of propagation based methods on all 'X\&Non-X' type datasets suggest that the class with semantically diverse patterns can result in sparser embeddings and down-sampling this class leads to high performance. 
For non-propagation based methods, on the other hand, we observed a reversed pattern of which variant of downsample yields high performance as compared with that for propagation based methods. This suggests that it is hard to predict which class will appear sparser in the embedding space  by solely knowing the data semantics, as it also depends on the embedding method itself.

Considering the importance of GNN based methods and its connection to propagation based methods, we carefully conducted a measurement study for certain propagation-based graph representation learning methods, namely Weisfeiler-Leman (WL) and propagation kernel (PK), and found the interaction between the mentioned factors and the sparsification property---where kernel similarity
of two graphs decrease with increasing number of propagations. This property leads to amplified density disparity, as the originally sparser class of graphs sparsify at a relatively higher rate. This is either a blessing or a curse for detection, depending on which scenario is considered as a testbed. When class with diverse patterns is down-sampled, sparsification helps disperse the outliers further, making the task easier. Otherwise, it makes the inliers look more dispersed. Sharing the similar underlying mechanism, OCGIN, the one-class end-to-end GNN based oulier detector we developed, has similar performance behavior on all 'X\&Non-X' datasets with WL and PK. However being an end-to-end trainable method, OCGIN has the ability of learning its parameters based mostly on the majority class, and surprisingly eliminates performance flip issue on all 'X\&Y' datasets.

Based on all our observations and analysis, we remark three important questions regarding GLOD as fundamental problems in this area, from three different perspectives: evaluation, model selection, and graph embedding method. Given the widely observed performance flip issue across models, traditional evaluation of models that simply averages performance across all datasets becomes problematic, as including worse-than-random performance does not appear to be the right thing. To avoid the worse-than-random scenario for a given new dataset, one may argue that doing model selection from among all candidate models to pick up one model without worse-than-random performance would be a good way out in practice. But one should also consider the hardness of unsupervised model selection and the high risk of picking up the wrong model with poor performance. Another way to avoid evaluation under performance flip is to design better graph embedding methods that can surpass the performance flip issue. 

\subsection{Discussion}

The three problems we outlined are essential but hard to answer. We present a discussion of our arguments in the following.  

Considering the performance flip is persistent and hard to avoid, caution should be taken when evaluating new model(s) under performance flip. Based on Hawkins's definition of outliers \cite{hawkins1980identification} -- that the outliers are simply the minority instances that are generated by a different mechanism -- no matter how we down-sample, defining outlier as the down-sampled class is meaningful. Then, evaluation by averaging performance across all datasets (including all variants of downsample) should be used. However, when worse-than-random performances are included, which skews the average performance, a model that appears competitive may result in considerably poor performance for certain cases. Another way to define the outlier class is based on dataset class semantics, where one class may be seen as a natural outlier class in the real world. However not all datasets have semantic outliers, and sometimes which class is outlier may be subjective and context-dependent (like in dataset AIDS, being active against HIV could be deemed as outlier since HIV is rare and semantically important, yet in the context of developing new drugs for treating AIDS, inactive drugs could be treated as the outliers). 

A way to sidestep performance flip altogether is to design better graph embedding methods that can overcome density disparity and overlapping support.  One can imagine that if the unsupervised embedding method can generate non-overlapping and clustered embeddings for each class, then performance flip would not arise. In other words, two-stage models are guaranteed to perform well for outlier detection if their graph embedding method can achieve good performance in graph-level clustering task. Another promising direction is to design  effective end-to-end models that can generate clustered representation for the majority class, by making use of the natural property of learning. However the current end-to-end graph representation learning models are all message-passing based graph neural networks which suffer from the inductive bias shared by WL and PK. 

In the prevalence of performance flip issue across existing models, it is not clear how to do fair and accurate model evaluation so as to pinpoint competitive models for GLOD. Assuming it is also hard to design new graph-level embedding models that can bypass performance flip no matter what dataset they are presented with, an open question remains as how to do GLOD in practice, i.e. in the wild, when presented with a new GLOD task. The answer is not trivial. Model evaluation and benchmarking is done exactly to identify a few competitive models to employ in practice. When evaluation has issues we have identified, a way out appears to do model selection. That is, instead of employing the ``best'' model based on  average performance across many historical/benchmark datasets, the goal would become designing an effective model selection strategy to choose a model to employ from a pool of existing models for a new task. 
Unfortunately, however, unsupervised model selection for GLOD is notoriously difficult  in the absence of any labels \cite{zhao2020automating}, and is perhaps an equally hard, if not harder, problem than grappling with performance flip and model evaluation (where labels are available for benchmark datasets, for evaluation). As with performance flip, it incurs the risk of selecting a poor model.  An easier yet practical direction may be to do model selection using (only) a small amount of labeled data, which however, has its own issues such as the representativeness of outliers in a small sample.

\subsection{Future work}

Several future directions that immediately span out of this study are three. First, whether our findings regarding the sparsification property of certain propagation based models have negative implications on (graph-level) clustering and classification tasks warrants further research. 
Second, whether our findings transfer to non-graph settings requires future analysis. We find that there is historical evidence that they do.\footnote{Arrhythmia, a (point-cloud) binary classification dataset from UCI Machine Learning Repository has been repurposed for outlier mining, by downsampling either one of the classes,  \url{http://homepage.tudelft.nl/n9d04/occ/514/oc_514.html} versus \url{http://homepage.tudelft.nl/n9d04/occ/515/oc_515.html}, respectively inducing scattered versus clustered outliers. Notice that the performance of the detection models are significantly higher for the former scenario.} Finally, our study can be extended to the analysis of class-level bias in model errors when multi-class classification datasets (with greater than two classes) are
used for outlier benchmark creation.

\section*{Acknowledgments}
This research is sponsored by NSF CAREER 1452425. We thank Charu Aggarwal for helpful discussions and critical reading of the manuscript. Code was partly built during the first author's internship at IBM T. J. Watson Research Center. Conclusions expressed in this material are those of the authors and do not necessarily reflect the views, expressed or implied, of the funding and other parties involved.

\bibliographystyle{ACM-Reference-Format}
\bibliography{reference}

\end{document}